\documentclass[fleqn,10pt,letterpaper]{article}
\usepackage[utf8]{inputenc}
\usepackage[T1]{fontenc}
\usepackage[english]{babel}
\usepackage{mathptmx}
\usepackage{helvet}
\usepackage{courier}
\usepackage{microtype}
\usepackage[left=2cm,right=2cm,top=2.25cm,bottom=2.25cm,headheight=12pt,letterpaper]{geometry}
\usepackage{amsmath,amsfonts,amssymb}
\usepackage{graphicx,xcolor}
\usepackage{booktabs}
\usepackage{authblk}
\usepackage{makecell}
\usepackage{footnote}
\usepackage[symbol]{footmisc}
\usepackage{textcomp}
\usepackage[superscript,biblabel,nomove]{cite}
\usepackage{jabbrv}
\usepackage{hyperref}
\hypersetup{colorlinks=true,allcolors=blue}
\date{}
\title{Million-scale multimodal pollen microscopy with expert-guided foundation models}

\author[1,*]{András Biricz}
\author[2]{Björn Gedda}
\author[3]{Donát Magyar}
\author[4]{Antonio Spanu}
\author[5]{János Fillinger}
\author[6,7,*,\textsuperscript{\textdagger}]{Péter Pollner}
\author[1,\textsuperscript{\textdagger}]{István Csabai}

\affil[1]{Department of Physics of Complex Systems, ELTE Eötvös Loránd University, Budapest, Hungary}
\affil[2]{The Palynological Laboratory at the Swedish Museum of Natural History, Stockholm, Sweden}
\affil[3]{National Centre for Public Health and Pharmacy, Budapest, Hungary}
\affil[4]{INRAE, UR 546 BioSP, Site Agroparc, Avignon 84914, France}
\affil[5]{National Korányi Institute for Pulmonology, Budapest, Hungary}
\affil[6]{Health Data Science and AI Knowledge Centre, Health Services Management Training Centre, Faculty of Health and Public Administration, Semmelweis University, Budapest, Hungary}
\affil[7]{Department of Biological Physics, ELTE Eötvös Loránd University, Budapest, Hungary}

\affil[ *]{Correspondence and requests for materials should be addressed to \textit{andras.biricz@ttk.elte.hu} or \textit{pollner@emk.semmelweis.hu}.}
\affil[ ]{\textsuperscript{\textdagger}Contributed equally to this work.}

\begin{document}

\flushbottom
\maketitle

\thispagestyle{empty}

\begin{abstract}
Automated pollen identification from microscopy remains a bottleneck in aerobiology, palaeoecology and biodiversity monitoring, because scalable systems must generalise across specimen preparation, scanner settings and geographic origins while retaining palynological interpretability. To address this gap, we present a million-scale multimodal pollen microscopy resource, Pollen AI Atlas, assembled from pure-species whole-slide bright-field images spanning four geographic origins, four scanner settings, 45 genera, and one family-only taxon across 31 botanical families. Seeded by one manually selected exemplar per source slide, token-level mining and filtering produced 1,511,390 released grain detections with 99.6\% proposal precision in expert-curated test regions. Each detection was paired with machine-generated grain-level morphological captions from five open-weight vision--language models, guided by expert-verified palynological anchors, yielding structured descriptions of aperture systems, wall ornamentation, shape and size. Among the evaluated models, Gemma4 provided the most controlled primary caption set, combining tight length control, no detected taxon-name or numeric-size leakage and the strongest text-retrieval performance. Baseline benchmarks with frozen visual features reached 88.16\% top-1 accuracy, while cross-regional retrieval showed that caption-derived text embeddings remained robust when image similarity degraded (mAP@20 0.811 versus 0.262). Released data, annotations, captions, splits, code, and weights provide a benchmark for pollen recognition, cross-regional domain adaptation and domain-specific multimodal microscopy learning.
\end{abstract}

\section*{Introduction}

Pollen analysis supports aeroallergen forecasting\cite{eHealth_Bastl2017}, palaeoclimate reconstruction\cite{Bourel2020PaleoPollenCNN}, and pollination ecology\cite{pollination_ecology_Bertrand}. In routine practice, however, microscopic identification still depends on trained specialists, and differences between analysts remain a recognised source of variability in standardised monitoring workflows\cite{Manual_standards_Galan2014}. This dependence limits throughput and standardisation at a time when long-running monitoring programmes have accumulated extensive microscopy archives and when climate-driven changes in pollen exposure are increasing the demand for timely and reproducible analysis\cite{french_hirst_history_article, swedish_hirst_history_article, DAmato2007_AllergenicPollen, climate_change_Barnes2018, Clot2024}. Although automatic and semi-automatic sensing approaches are increasingly available, digitised bright-field microscopy of Hirst-type samples\cite{HIRST_pollen_trap} remains among the most widely deployed and practically accessible modalities in operational aerobiology\cite{Viertel2022PollenSurvey, Principles_meths_for_automated_pollen}. Pure-species reference slides provide an efficient bridge between this operational microscopy context and machine-learning dataset construction\cite{closematchpaper_semi_automated_annots, Biricz2025}. Because each slide contains pollen from a single identified taxon while remaining compatible with bright-field preparation and slide-scanning workflows, slide-level taxonomic labels can serve as weak supervision without exhaustive per-grain annotation. Such collections are therefore not substitutes for mixed environmental samples, but they are an efficient and scientifically grounded source for building reusable, expert-grounded corpora.

Public pollen image resources have substantially advanced computational pollen analysis, but existing datasets remain constrained by at least one of taxonomic breadth, sample count, imaging modality, or source diversity. These resources include morphological reference databases such as PalDat\cite{PalDat2000} and image collections such as Pollen13K\cite{Battiato2020Pollen13K}, POLLEN73S\cite{Astolfi2020Pollen73S}, POLEN23E\cite{Goncalves2016PollenML}, and POLLEN24\_SP\cite{Valiente2025MonofloralHoney}. They have supported classification studies in bright-field and related microscopy settings\cite{Sevillano2018POLEN23E, Sevillano2020Pollen46, manualslide_prep_cnn_Polling2021}, but most remain image-only, application-specific, or limited in source diversity. Consequently, cross-dataset and cross-region generalisation remains a persistent challenge in automated pollen analysis\cite{robustdet_cnn_example, Viertel2022PollenSurvey, Principles_meths_for_automated_pollen}. Recent work has begun to integrate pollen images with structured morphological attributes, demonstrating the value of explicit palynological information, but such resources remain small relative to whole-slide archives and provide tabulated attributes rather than automated per-grain natural-language records\cite{Zhang2026PollenAttributes}. Palynology therefore still lacks a resource that is simultaneously large-scale, acquisition-diverse, expert-grounded, and paired at grain level with natural-language morphological records.

At the same time, the technical basis for dataset construction has changed. Our earlier detector-oriented study established the feasibility of this weak-supervision strategy in a conventional object-detection setting, using single-exemplar OWL-ViT initialisation, YOLOS-based refinement, and Faster R-CNN training across seven taxa and three regions\cite{Biricz2025, owl_objdet_model_1_2022}. In the present work, the same weak-supervision principle is redirected from generating detector training annotations to mining dense visual representations directly at whole-slide scale. Conventional closed-set detectors can be brittle when deployed on taxa, preparations, scanners, or staining conditions that differ from the labelled training regime\cite{Dhamija2020TheOE, Joseph2021TowardsOW}, motivating alternatives that use reusable dense visual representations rather than task-specific box-regression models. Self-supervised vision transformers provide one such alternative. DINO revealed emergent object segmentation in ViT attention maps\cite{Caron_2021_ICCV}; subsequent work showed that ViT descriptors encode fine-grained semantic parts\cite{amir2021deep}; and DINOv2 established dense self-supervised features as robust representations transferable across tasks\cite{Oquab2023DINOv2LR}. Combined with promptable segmentation models such as SAM~2\cite{Ravi2024SAM2S}, these developments make token-level search a plausible route for microscopy. Rather than training a detector to enumerate objects, a manually selected exemplar from each source slide can seed dense feature-space mining, with segmentation and filtering applied only to candidate grains. The present work tests whether this route can yield a high-precision pollen corpus at whole-slide scale.

A second shift concerns language. Pollen morphology is already expressed through standardised descriptive terminology covering aperture systems, exine ornamentation, polarity, shape, and size\cite{Pollen_book_terminology, beug2021leitfaden, terminology_paper_pollen}. This makes palynology unusually well suited to vision--language modelling, provided that generation is constrained by expert-curated morphological anchors rather than treated as unconstrained image captioning. In neighbouring microscopy fields, especially digital pathology, multimodal foundation models now use image--text supervision, diagnostic reports, and generated morphological descriptions for retrieval, classification, and whole-slide learning\cite{conch_mahmoodlab, pathchat_mahmoodlab, titan_mahmoodlab}. Broader microscopy benchmarks likewise show that image--language alignment is becoming a central problem across biological imaging domains\cite{Lozano_2025_CVPR, MicroBench2024}. Palynology has not yet had an equivalent large-scale image--caption resource linking individual bright-field pollen crops to structured morphological language and slide-level taxonomic context.

Here we present Pollen AI Atlas, a million-scale multimodal pollen microscopy resource assembled from pure-species whole-slide bright-field images spanning multiple geographic collections, scanner settings, 45 genera, and one family-only taxon across 31 botanical families. The atlas combines single-exemplar token-level mining, precision-oriented filtering, expert-curated reference annotations, and expert-anchored structured morphological captions generated by five recent open-weight vision--language models (VLMs). Caption generation is guided by expert-verified palynological anchors derived primarily from Beug's \textit{Leitfaden der Pollenbestimmung}\cite{beug2021leitfaden} and supplemented by PalDat\cite{PalDat2000}. This design addresses three research questions: whether one manually selected exemplar per source slide can scale pollen-grain discovery across whole-slide microscopy while preserving the precision required for a scientific reference corpus; whether expert-anchored VLMs can convert image-only crops into controlled, taxon-discriminative morphological records; and whether the resulting multimodal corpus supports supervised classification and cross-regional retrieval across scanner, slide-source, and geographic shifts. To our knowledge, Pollen AI Atlas is the first million-scale pollen image--caption microscopy resource pairing individual bright-field pollen grain crops with structured natural-language morphological descriptions and expert-verified slide-level taxonomic labels. The atlas is intended as a high-precision, expert-guided, pure-species reference resource for pollen classification, retrieval, and domain-specific vision--language modelling, not as an exhaustive detector for arbitrary mixed environmental samples.

\section*{Results}

\begin{figure}[ht!]
\centering
\includegraphics[width=0.903\textwidth]{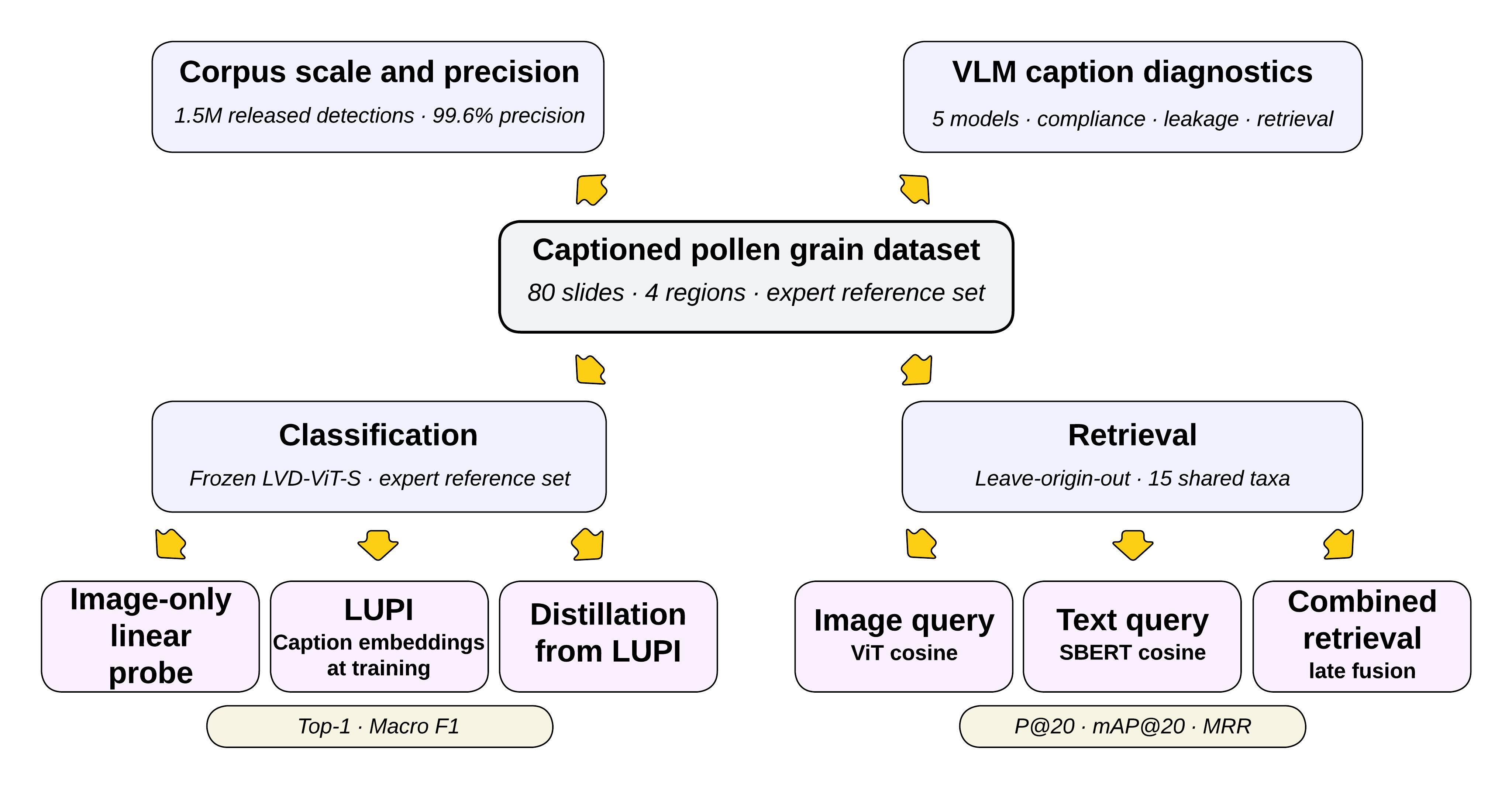}
\caption{The released captioned pollen-grain dataset is evaluated through complementary corpus-level, caption-level and downstream benchmarks. Corpus assembly is assessed by the number of released detections and expert-curated detection precision. Caption generation is assessed through five-model VLM diagnostics, including prompt compliance, leakage control, vocabulary coverage and text-retrieval performance. Downstream classification uses original frozen LVD-ViT-S image features for image-only linear probing, privileged-information training with caption embeddings, and distillation from the privileged-information teacher. Downstream retrieval compares image, text and late-fusion search under leave-origin-out evaluation across 15 shared genus-level targets. Classification reports Top-1 accuracy and Macro-F1; retrieval reports P@1, P@20, mAP@20 and MRR.}
\label{fig:evaluation_overview}
\end{figure}

Results are organised around the three research questions introduced above: corpus-scale grain discovery, controlled morphological captioning, and downstream classification and retrieval utility. Fig.~\ref{fig:evaluation_overview} summarises this structure by linking corpus-scale validation, five-model caption diagnostics, supervised classification and cross-regional retrieval within the released captioned pollen-grain dataset.

\subsection*{Corpus scale and precision}

The dataset assembly pipeline was applied to 85 whole-slide bright-field microscopy images spanning four geographic origins, four scanner settings, and mixed magnifications (Supplementary Table~\ref{tab:dataset_summary}). For each source slide, a manually selected exemplar grain seeded token-level mining across the whole-slide feature space, yielding 1,826,529 candidate objects. Multi-stage quality filtering by attention-based reranking, prototype refinement, non-maximum suppression, and classifier gating retained 1,528,984 detections, corresponding to an overall retention of 83.7\%. After exclusion of five slides for taxonomic or image-quality reasons, the release corpus comprised 1,511,390 filtered detections from 80 slides, covering 45 genera and one family-only taxon across 31 botanical families.

Detection quality was evaluated against Test Set 2 (TS2), the primary expert-curated evaluation set. Its quantitative subset contained 10,335 pollen annotations from 80 slides, comprising 79 slides with expert-verified taxon identifications at genus level or finer and one captioned slide identified only to the family Fabaceae, for which the genus-level label was recorded as \textit{Unknown}. Within these regions, pipeline proposals preserved by the expert were counted as true positives, deleted proposals as false positives, and expert-added grains as false negatives. Across the 80 evaluated slides, the filtered pipeline produced 4,506 true positives, 20 false positives, and 5,829 false negatives, corresponding to 99.6\% precision, 43.6\% recall, and an F1 score of 60.6\% (Table~\ref{tab:detection_quality}). Precision exceeded 99\% in every regional subset, ranging from 99.0\% in the Mediterranean collection to 100.0\% in the Hungarian collection.

\begin{table}[ht!]
\centering
\begin{tabular}{lrrrrrr}
\toprule
Dataset & TP & FP & FN & Precision & Recall & F1 \\
\midrule
French & 2,567 & 9 & 2,011 & 99.7\% & 56.1\% & 71.8\% \\
Hungarian & 414 & 0 & 967 & 100.0\% & 30.0\% & 46.1\% \\
Mediterranean & 510 & 5 & 740 & 99.0\% & 40.8\% & 57.8\% \\
Swedish & 1,015 & 6 & 2,111 & 99.4\% & 32.5\% & 49.0\% \\
\midrule
\textbf{Overall} & \textbf{4,506} & \textbf{20} & \textbf{5,829} & \textbf{99.6\%} & \textbf{43.6\%} & \textbf{60.6\%} \\
\bottomrule
\end{tabular}
\caption{Detection quality against expert-curated test annotations. Pipeline detections were evaluated against the TS2 expert ground truth in one 5,000$\times$5,000\,px test region per slide. True positives (TP), false positives (FP) and false negatives (FN) were accumulated by dataset origin; precision, recall and F1 were computed as TP/(TP+FP), TP/(TP+FN) and $2PR/(P+R)$, respectively. The evaluation characterises the high-precision operating point used for atlas construction rather than an exhaustive-detection setting.}
\label{tab:detection_quality}
\end{table}

Slide-level results showed that the high precision was not driven by a small number of high-yield or visually favourable slides. Mean slide-level precision was 99.0\%, median precision was 100.0\%, and 68 of 80 slides achieved 100.0\% precision; only three slides fell below 95\%. Recall was lower and more variable, consistent with the intended precision-first operating point of the workflow. Per-slide detection metrics are provided in Supplementary Table~\ref{tab:detection_per_slide}. At corpus scale, this operating point provides broad per-taxon coverage while keeping the measured false-positive rate below 0.5\% in the evaluated regions.

\subsection*{Expert-anchored VLM captions and automated caption diagnostics}

The 1,511,390 filtered pollen grains in the release corpus were captioned independently by five off-the-shelf VLMs: Qwen2.5-VL\cite{bai2025qwen25vltechnicalreport}, Qwen3-VL\cite{bai2025qwen3vltechnicalreport}, Qwen3.5\cite{QWEN35_27B}, Qwen3.6\cite{QWEN36_27B}, and Gemma4\cite{gemma4_31b_it}. Each model received the same slide-specific instruction set, and the full five-model run produced 7,556,950 grain-level captions without VLM fine-tuning.

\begin{table*}[ht]
\centering
\small
\begin{tabular}{lccccc}
\toprule
Metric & Gemma4 & Qwen2.5-VL & Qwen3-VL & Qwen3.5 & Qwen3.6 \\
\midrule
Word count, mean $\pm$ s.d. & 70.6 $\pm$ 2.0 & 137.9 $\pm$ 18.4 & 89.7 $\pm$ 22.0 & 76.8 $\pm$ 8.9 & 66.8 $\pm$ 9.6 \\
Taxon-name leakage & 0.0\% & 0.0\% & 0.0\% & 0.0001\% & 0.0\% \\
Numeric size leakage & 0.0\% & 42.5\% & 0.2\% & 0.1\% & 0.0\% \\
Debris-screening opener & 7.21\% & 0.06\% & 76.65\% & 0.44\% & 4.53\% \\
Anchor-vocabulary coverage & 93.6\% (395/422) & 98.8\% (417/422) & 94.5\% (399/422) & 97.4\% (411/422) & 97.9\% (413/422) \\
\midrule
ALL text retrieval P@1 & \textbf{0.867 [0.67, 1.00]} & 0.200 [0.00, 0.40] & 0.400 [0.13, 0.67] & 0.600 [0.33, 0.87] & 0.733 [0.53, 0.93] \\
ALL text retrieval mAP@20 & \textbf{0.834 [0.69, 0.96]} & 0.207 [0.04, 0.40] & 0.279 [0.11, 0.48] & 0.350 [0.20, 0.52] & 0.718 [0.51, 0.90] \\
ALL text retrieval MRR & \textbf{0.890 [0.73, 1.00]} & 0.346 [0.17, 0.54] & 0.435 [0.21, 0.68] & 0.700 [0.50, 0.89] & 0.788 [0.59, 0.95] \\
\bottomrule
\end{tabular}
\caption{Automated caption diagnostics and ALL-mode text-retrieval comparison for the five VLM caption sets. The upper block reports caption-control diagnostics computed separately for each model-specific caption set ($n=1,511,390$ captions per model), including caption length, taxon-name leakage, numeric size leakage, debris-screening opener frequency, and anchor-vocabulary coverage. Values reported as mean $\pm$ s.d. use the standard deviation across captions. Anchor-vocabulary coverage is the fraction of the 422-term anchor-derived palynological vocabulary observed at least once in each caption set. The debris-screening opener rate is retained as a prompt-style diagnostic and should not be interpreted as expert-labelled debris prevalence. The lower block reports ALL-mode text retrieval against each model-specific 1,480,653-caption genus-resolved index without origin exclusion, using SBERT all-MiniLM-L6-v2 embeddings of independently rephrased expert descriptors as one text query for each of 15 genus-level targets. Metrics give equal weight to each target; 95\% confidence intervals are from 10,000 target-level bootstrap resamples using the percentile method. Best operational retrieval metrics are shown in bold.}
\label{tab:caption_diagnostics}
\end{table*}

Automated diagnostics showed strong control over the most important data-leakage constraint: explicit taxon-name leakage was absent or vanishingly rare across the five caption sets (Table~\ref{tab:caption_diagnostics}). The main differences between models were in length control and numerical-measurement leakage. Gemma4 produced the most tightly constrained outputs, with a mean caption length of 70.6$\pm$2.0 words and all captions within the requested 60--80 word range. Qwen2.5-VL generated substantially longer captions (137.9$\pm$18.4 words) and reported numeric grain dimensions in 42.5\% of outputs despite instructions to avoid numeric measurements. The later Qwen models and Gemma4 showed minimal numeric size leakage (0.0--0.2\%). Corpus-level anchor-vocabulary coverage ranged from 93.6\% to 98.8\% of the 422-term lexicon. Vocabulary breadth did not translate monotonically into retrieval quality: Qwen2.5-VL covered 98.8\% of the vocabulary but reached mAP@20 = 0.207, whereas Gemma4 covered 93.6\% but reached mAP@20 = 0.834. This suggests that concise, controlled use of diagnostic terminology preserved embedding-level semantics better than broad term use alone.

To test whether the generated text supported semantic retrieval, independently rephrased expert descriptors were embedded as text queries and compared with each 1,480,653-caption index restricted to the 45 genus-level labels in ALL mode, which retained the complete index without origin exclusion. Gemma4 produced the strongest operational text-retrieval scores, with P@1 = 0.867, mAP@20 = 0.834, and MRR = 0.890. Qwen3.6 was the closest comparator (P@1 = 0.733, mAP@20 = 0.718, MRR = 0.788), whereas Qwen2.5-VL, Qwen3-VL and Qwen3.5 produced lower mAP@20 values (0.207--0.350). These diagnostics do not by themselves establish per-grain expert-level morphological correctness; rather, they show that the captioning protocol generated controlled, domain-terminology-rich, genus-discriminative text that supports semantic search under a common embedding and query setup. On this basis, Gemma4 was selected as the primary caption set for downstream utility testing, while the remaining four caption sets were retained as comparative outputs.

\subsection*{Classification baseline and domain shift}

To establish a supervised baseline, a linear classification head was trained on frozen features from the original LVD-ViT-S backbone to assess how well the released image records support genus-level pollen identification. The evaluation set comprised 10,218 valid test samples across 45 genus-level target labels from the 10,335-grain expert-curated reference set, after excluding records from the Fabaceae slide for which the genus-level label was \textit{Unknown}. Because TS2 was generated by correcting pipeline proposals rather than accepting retained detections alone, this benchmark includes both confirmed pipeline detections and expert-added missed grains from manually corrected test regions. Each configuration was repeated with five random seeds, and results are reported as mean$\pm$SE.

\begin{table}[ht!]
\centering
\small
\begin{tabular}{@{}rl l r@{\,}c@{\,}l r@{\,}c@{\,}l r@{}}
\toprule
Exp & Setting & Region & \multicolumn{3}{c}{Top-1 (\%)} & \multicolumn{3}{c}{Macro-F1 (\%)} & $N$ \\
\midrule
\multicolumn{10}{l}{\textit{Image-only linear probe}} \\
1 & Linear probe & All & 88.16 & $\pm$ & 0.15 & 85.24 & $\pm$ & 0.20 & 10,218 \\
10 & \quad w/ stain norm. & All & 66.96 & $\pm$ & 0.36 & 64.53 & $\pm$ & 0.27 & 10,218 \\
\cmidrule(lr){1-10}
2 & Linear probe & French & 71.26 & $\pm$ & 0.53 & 69.83 & $\pm$ & 0.30 & 9,950 \\
6 & \quad w/ stain norm. & French & 49.62 & $\pm$ & 0.56 & 53.68 & $\pm$ & 0.41 & 9,950 \\
\cmidrule(lr){1-10}
3 & Linear probe & Hungarian & 74.92 & $\pm$ & 0.32 & 64.71 & $\pm$ & 0.37 & 2,602 \\
7 & \quad w/ stain norm. & Hungarian & 85.57 & $\pm$ & 0.14 & 81.16 & $\pm$ & 0.69 & 2,602 \\
\cmidrule(lr){1-10}
4 & Linear probe & Swedish & 69.93 & $\pm$ & 0.20 & 63.05 & $\pm$ & 0.15 & 6,433 \\
8 & \quad w/ stain norm. & Swedish & 58.78 & $\pm$ & 0.32 & 44.95 & $\pm$ & 0.17 & 6,433 \\
\cmidrule(lr){1-10}
5 & Linear probe & Mediterranean & 66.16 & $\pm$ & 0.58 & 60.98 & $\pm$ & 0.63 & 3,927 \\
9 & \quad w/ stain norm. & Mediterranean & 56.09 & $\pm$ & 0.07 & 34.85 & $\pm$ & 0.52 & 3,927 \\
\midrule
\multicolumn{10}{l}{\textit{Caption-informed training settings}} \\
11 & LUPI & All & 87.45 & $\pm$ & 0.12 & 84.51 & $\pm$ & 0.12 & 10,218 \\
16 & Distillation & All & 88.30 & $\pm$ & 0.10 & 85.53 & $\pm$ & 0.15 & 10,218 \\
17 & Distillation & French & 71.25 & $\pm$ & 0.23 & 70.57 & $\pm$ & 0.13 & 9,950 \\
18 & Distillation & Hungarian & 74.80 & $\pm$ & 0.21 & 64.59 & $\pm$ & 0.27 & 2,602 \\
19 & Distillation & Swedish & 69.59 & $\pm$ & 0.16 & 62.85 & $\pm$ & 0.16 & 6,433 \\
20 & Distillation & Mediterranean & 65.61 & $\pm$ & 0.57 & 59.80 & $\pm$ & 0.42 & 3,927 \\
\bottomrule
\end{tabular}
\caption{Baseline classification experiments on released pollen-grain image records. All experiments used original ViT-Small-LVD image features and no backbone fine-tuning. Values report mean\,$\pm$\,standard error across five random seeds (41--45). TS2 denotes the primary expert-curated test set. The \textit{Unknown} label was excluded, yielding 45 genus-level target labels and 10,218 valid test samples for the combined setting. $N$ gives the number of TS2 samples evaluated in each row. For regional training settings, evaluation includes all TS2 grains whose genus-level target labels were present in the corresponding training subset, irrespective of geographic origin, thereby testing cross-regional transfer implicitly. Stain-normalised rows use Macenko normalisation\cite{Macenko2009StainNorm}. Caption-informed settings include learning using privileged information (LUPI), where SBERT caption embeddings\cite{Reimers2019SentenceBERTSE} are available during training, and image-only student models distilled from LUPI teachers.}
\label{tab:classification}
\end{table}

When trained on all four geographic origins jointly, the image-only linear probe achieved 88.16$\pm$0.15\% Top-1 accuracy and 85.24$\pm$0.20\% Macro-F1 (Table~\ref{tab:classification}). Models trained on a single origin and evaluated on eligible evaluation samples whose genus-level target labels were present in the corresponding training subset achieved lower aggregate Top-1 accuracy, ranging from 66.16\% for Mediterranean-trained data to 74.92\% for Hungarian-trained data. These results indicate that training on a single source collection does not cover the full acquisition and taxonomic variability present in the atlas.

Macenko stain normalisation was evaluated as a traditional single-modality image domain-adaptation baseline. It reduced performance for the combined-origin model, from 88.16\% to 66.96\% Top-1 accuracy (-21.2 percentage points), and also degraded the French, Swedish, and Mediterranean regional models. The exception was the Hungarian model, where stain normalisation increased Top-1 accuracy from 74.92\% to 85.57\% (+10.7 percentage points). This pattern indicates that this single image-only normalisation strategy was not uniformly beneficial across the heterogeneous scanner, preparation, and staining conditions represented in the atlas. Because collection identity, staining protocol, slide age or fading, scanner response and taxon composition covary in these regional subsets, this result should not be interpreted as an isolated ageing or staining effect; rather, it shows that a single colour-normalisation transform can help one acquisition regime while harming others.

Two caption-informed training settings were then evaluated using SBERT embeddings of the Gemma4 captions. Learning using privileged information (LUPI) concatenated image and text embeddings during training but used image features only at inference; this yielded 87.45$\pm$0.12\% Top-1 accuracy, slightly below the image-only baseline. Distillation from a LUPI teacher to an image-only student reached 88.30$\pm$0.10\% Top-1 accuracy and 85.53$\pm$0.15\% Macro-F1, the highest point estimate in the experiment suite, but the gain over the image-only baseline was small (+0.14 percentage points). Per-region distillation models did not materially improve over their corresponding image-only regional baselines. Across these genus-level linear-probe experiments, Gemma4 caption embeddings did not materially change classification accuracy relative to the frozen visual features; the retrieval experiments below evaluate the caption modality more directly.

\subsection*{Cross-regional retrieval}

We evaluated semantic retrieval with the Gemma4-captioned corpus using the ALL and CROSS-REG protocols. After retrieval-specific embedding matching, the search index contained 1,480,653 grains from 79 slides (see Methods for exclusion details). ALL mode, introduced above as the full-corpus ceiling condition, used the complete index without geographic exclusion. CROSS-REG mode excluded all grains from the query's geographic origin, providing a leave-origin-out test across scanners, preparation protocols, staining conditions, and source collections. The query set was restricted to 15 genus-level targets represented in at least two geographic origins. Text queries used independently rephrased expert morphological descriptors, and image queries used exemplar crops.

\begin{table}[ht!]
\centering
\small
\begin{tabular}{llrrrr}
\toprule
Mode & Modality & P@1 & P@20 & mAP@20 & MRR \\
\midrule
ALL & Image & \makecell{1.000\\{\footnotesize [1.00,\,1.00]}} & \makecell{0.939\\{\footnotesize [0.86,\,0.99]}} & \makecell{0.917\\{\footnotesize [0.83,\,0.98]}} & \makecell{1.000\\{\footnotesize [1.00,\,1.00]}} \\
ALL & Text & \makecell{0.867\\{\footnotesize [0.67,\,1.00]}} & \makecell{0.897\\{\footnotesize [0.80,\,0.98]}} & \makecell{0.834\\{\footnotesize [0.69,\,0.96]}} & \makecell{0.890\\{\footnotesize [0.73,\,1.00]}} \\
ALL & Combined & \makecell{0.993\\{\footnotesize [0.96,\,1.00]}} & \makecell{0.977\\{\footnotesize [0.93,\,1.00]}} & \makecell{0.970\\{\footnotesize [0.92,\,1.00]}} & \makecell{0.995\\{\footnotesize [0.98,\,1.00]}} \\
\midrule
CROSS-REG & Image & \makecell{0.311\\{\footnotesize [0.14,\,0.49]}} & \makecell{0.304\\{\footnotesize [0.15,\,0.47]}} & \makecell{0.262\\{\footnotesize [0.12,\,0.42]}} & \makecell{0.379\\{\footnotesize [0.21,\,0.55]}} \\
CROSS-REG & Text & \makecell{0.867\\{\footnotesize [0.69,\,1.00]}} & \makecell{0.871\\{\footnotesize [0.75,\,0.97]}} & \makecell{0.811\\{\footnotesize [0.65,\,0.94]}} & \makecell{0.901\\{\footnotesize [0.77,\,1.00]}} \\
CROSS-REG & Combined & \makecell{0.644\\{\footnotesize [0.41,\,0.86]}} & \makecell{0.631\\{\footnotesize [0.43,\,0.81]}} & \makecell{0.593\\{\footnotesize [0.38,\,0.78]}} & \makecell{0.662\\{\footnotesize [0.44,\,0.86]}} \\
\midrule
\multicolumn{6}{l}{\textit{Label-shuffled negative controls, averaged across three seeds}} \\
\midrule
ALL & Image & \makecell{0.052\\{\footnotesize [0.00,\,0.15]}} & \makecell{0.046\\{\footnotesize [0.02,\,0.08]}} & \makecell{0.012\\{\footnotesize [0.00,\,0.02]}} & \makecell{0.132\\{\footnotesize [0.05,\,0.24]}} \\
ALL & Text & \makecell{0.044\\{\footnotesize [0.00,\,0.11]}} & \makecell{0.047\\{\footnotesize [0.02,\,0.08]}} & \makecell{0.011\\{\footnotesize [0.00,\,0.02]}} & \makecell{0.123\\{\footnotesize [0.04,\,0.23]}} \\
CROSS-REG & Image & \makecell{0.063\\{\footnotesize [0.00,\,0.18]}} & \makecell{0.050\\{\footnotesize [0.02,\,0.09]}} & \makecell{0.014\\{\footnotesize [0.00,\,0.03]}} & \makecell{0.143\\{\footnotesize [0.06,\,0.26]}} \\
CROSS-REG & Text & \makecell{0.033\\{\footnotesize [0.00,\,0.08]}} & \makecell{0.041\\{\footnotesize [0.02,\,0.07]}} & \makecell{0.010\\{\footnotesize [0.00,\,0.02]}} & \makecell{0.118\\{\footnotesize [0.05,\,0.21]}} \\
\bottomrule
\end{tabular}
\caption{Gemma4 multimodal retrieval results. Retrieval was performed against a search index of 1,480,653 grains from 79 slides with matched image and Gemma4 caption embeddings. The query set comprised 15 genus-level targets represented in at least two geographic origins. ALL mode uses the full search index, whereas CROSS-REG excludes all grains from the query's geographic origin as a leave-origin-out test. Modalities are image retrieval using original frozen ViT-Small-LVD features, text retrieval using SBERT all-MiniLM-L6-v2 embeddings, and late-fusion retrieval with equal image--text weighting ($\alpha=0.5$). Text queries were expert-derived morphological descriptors independently rephrased from validated anchor descriptions, not raw prompt anchors. Metrics are macro-averaged per query genus-level target; P@1 and P@20 denote precision at rank 1 and 20, mAP@20 denotes mean average precision at 20, and MRR denotes mean reciprocal rank. Scores give equal weight to targets and origins. Brackets for unshuffled values show 95\% percentile intervals from 10,000 hierarchical bootstrap resamples. Label-shuffled controls are three-seed means, with lower and upper descriptive bounds averaged separately across seeds.}
\label{tab:retrieval_results}
\end{table}

In ALL mode, image retrieval reached P@1 = 1.000 and mAP@20 = 0.917, consistent with same-slide and same-origin visual similarity dominating the top ranks when no geographic exclusion was applied (Table~\ref{tab:retrieval_results}). Text retrieval was also strong, but lower than image retrieval, with P@1 = 0.867 and mAP@20 = 0.834. Late-fusion combined retrieval produced the highest ALL-mode mAP@20 (0.970), indicating that image and text similarities were complementary when same-origin visual cues remained available.

The pattern changed under CROSS-REG exclusion. Image retrieval dropped to P@1 = 0.311 and mAP@20 = 0.262, whereas text retrieval remained high, with P@1 = 0.867, mAP@20 = 0.811, and MRR = 0.901. Thus, the caption-derived morphological representation retained most of its ALL-mode retrieval performance after removal of the query origin, while image similarity did not. Combined retrieval occupied an intermediate position (mAP@20 = 0.593), consistent with the fixed $\alpha=0.5$ fusion weight averaging a strong cross-regional text signal with a weaker cross-regional image signal.

Label-shuffled controls were included to estimate chance-level retrieval under matched corpus composition. In the modality-specific controls, P@1 collapsed to 0.033--0.063 and mAP@20 to 0.010--0.014. Relative to these matched shuffled baselines, mAP@20 was 19--81 times higher in the main retrieval runs, with the strongest separation observed for CROSS-REG text retrieval. This separation indicates that the retrieval scores reflect taxon-level structure rather than label frequency or corpus composition, and that the strongest origin-excluded retrieval result came from the caption-derived text modality.

Supplementary Fig.~\ref*{fig:umap} provided a qualitative representation-level check consistent with the retrieval results. Image embeddings showed stronger origin-associated structure, whereas Gemma4 caption embeddings grouped same-taxon grains more consistently across geographic origins. This visual pattern supports the quantitative CROSS-REG result that morphology-focused text remained more robust than image similarity under scanner and preparation shifts.

\section*{Discussion}

The assembled pipeline and resulting Pollen AI Atlas show that million-scale multimodal pollen microscopy can be built from two deliberately small, expert-guided inputs: one visual exemplar per source slide for whole-slide image mining and human-in-the-loop verified morphological anchors for caption generation. Pure-species reference slides make this weak supervision possible by providing reliable slide-level taxonomic context while keeping manual effort concentrated where it is most informative. The atlas should therefore be read as a paired image--text reference corpus for training and benchmarking, not as a substitute for mixed environmental samples or exhaustive manual slide-level counting.

For the image modality, single-exemplar token-level mining and filtering produced a high-precision, million-scale pollen corpus with a measured false-positive rate below 0.5\% across heterogeneous slide preparations, scanner settings, and geographic origins. This answers the first research question in a precision-first sense: the atlas is not an exhaustive enumeration of every grain, but it shows that one expert-selected exemplar per source slide can seed large-scale discovery while preserving the purity needed for a reusable reference corpus. Earlier pollen datasets and classifier benchmarks have typically contained hundreds to tens of thousands of images rather than more than 1.5 million filtered grain records\cite{Goncalves2016PollenML, Battiato2020Pollen13K, Sevillano2020Pollen46, Zhang2026PollenAttributes}. Its scientific credibility comes from using expertise at the points where it most strongly determines data quality: exemplar selection, human-in-the-loop verification of positive and negative crops during backbone initialisation, pollen-specific fine-tuning of the visual encoder, and expert-curated ground truth. In this respect, the present study extends our earlier detector-focused workflow, which used OWL-ViT initialisation, refinement loops, and Faster R-CNN training for seven taxa across three regions, into a 46-label paired image--text atlas\cite{Biricz2025, owl_objdet_model_1_2022}.

The central methodological shift is that atlas construction is framed as retrieval in a dense feature space rather than as supervised detector training alone. YOLO- and Faster R-CNN-style pipelines remain strong within-domain baselines when sufficient labelled boxes are available, but they depend on box-regression training and a predefined target set, making them fragile when new taxa, slide preparations, or scanner conditions enter deployment\cite{Dhamija2020TheOE, Joseph2021TowardsOW, yolo_rcnn_example_Chaves2024_webview}. The present pipeline instead searches self-supervised ViT token features and applies SAM-based mask refinement only after candidate discovery. This design follows the DINO-to-DINOv2 trajectory toward reusable dense visual features and combines it with SAM~2 promptable spatial refinement\cite{Caron_2021_ICCV, amir2021deep, Oquab2023DINOv2LR, Ravi2024SAM2S}. Recent DINOv3 and SAM~3 developments reinforce the same direction: dense visual features and promptable concept segmentation are becoming increasingly reusable components for discovery by exemplars, concepts, or text prompts\cite{simeonidinov3, carion2026sam3segmentconcepts}. The contribution is therefore not that classical detectors are obsolete, but that expert-guided dense retrieval can scale atlas construction before task-specific detector training is needed.

The generated captions have a specific scientific status within the atlas. They are machine-generated morphological descriptions produced under expert-curated constraints; they should not be read as independent expert annotations or taxonomic ground truth. The anchors, drawn primarily from Beug's \textit{Leitfaden der Pollenbestimmung} with PalDat as a supplementary source, impose a standard palynological vocabulary for apertures, ornamentation, shape, polarity, and size while still allowing each VLM to describe the visible crop\cite{beug2021leitfaden, PalDat2000, Pollen_book_terminology}. This makes morphology searchable and comparable at scale. The five-model comparison shows that the task is controlled but not trivial: leakage was absent or vanishingly rare, vocabulary coverage was broad, and semantic retrieval identified meaningful model differences. Gemma4 was selected for downstream evaluation because it combined strong prompt compliance with the best full-corpus text-retrieval behaviour. Related multimodal pathology work similarly uses domain text, reports and generated descriptions as supervisory signals for representation learning and retrieval, while keeping them distinct from independently verified object-level labels\cite{conch_mahmoodlab, pathchat_mahmoodlab, titan_mahmoodlab}. Our contribution is to apply that image--language pattern to bright-field pollen crops at grain level, at million scale, and under explicit expert-guided morphology constraints.

The clearest multimodal finding is that morphology-focused text is more robust across scanner and preparation shifts than image similarity alone. This mirrors expert palynological practice in a simplified computational form: visible grains are interpreted through morphology such as apertures, ornamentation, outline, and polarity, rather than through acquisition-specific colour or illumination cues. In ALL mode, where same-origin visual cues remain available, image retrieval reached P@1 = 1.000 and mAP@20 = 0.917. In CROSS-REG mode, which removes all grains from the query's geographic origin, image retrieval dropped to P@1 = 0.311 and mAP@20 = 0.262, whereas Gemma4 text retrieval remained high at P@1 = 0.867 and mAP@20 = 0.811. A direct interpretation is that the expert-anchored caption layer suppresses acquisition-specific variation that remains encoded in image features, such as staining intensity, illumination, point-spread differences, mounting artefacts, and focus-fusion characteristics. Because the retrieval queries were independently phrased expert morphological descriptions rather than the same expert-verified palynological anchors used for captioning, this result should be read as evidence for a robust and useful text modality. The UMAP visualisation supports this interpretation descriptively. Because the late-fusion experiment used a fixed \(\alpha=0.5\) weight, the combined result should be read as a transparent baseline rather than as an optimised multimodal method.

Cross-regional generalisation remains the main practical challenge. The combined-origin frozen-feature linear probe reached 88.16\% Top-1 accuracy on the evaluation set, but single-origin transfer was uneven. Retrieval exposed the same issue from another angle: when same-origin matches were removed, image similarity became substantially less reliable, and transfer was asymmetric because source size, slide cleanliness, preparation quality, and taxon overlap differed by origin. Macenko normalisation improved the Hungarian regional model by 10.7 percentage points but reduced the combined-origin model by 21.2 percentage points, indicating that scanner and preparation variation is not a single removable colour shift\cite{Macenko2009StainNorm}. Similar transfer difficulties have been reported in automated pollen monitoring and palaeoecological pollen workflows, where models trained on clean or region-specific material do not necessarily transfer to new acquisition or preservation conditions\cite{Biricz2025, Bourel2020PaleoPollenCNN, Durand2024PollenAutoAnnot}. The atlas therefore provides a practical benchmark for testing which domain shifts can be normalised away and which require pollen-specific adaptation.

Several limitations follow directly from the design. The mining pipeline is precision-first, so recall is moderate and slide-dependent, varying with slide density, slide quality, query representativeness, and entropy-based stopping. The released detections should therefore not be used for quantitative slide-level counting, abundance estimation or within-taxon trait-frequency analyses without additional validation. Expert validation was focused on necessary quality-control points rather than applied exhaustively to every released grain, an inherent tradeoff at million-grain scale. The source material consists of pure-species reference slides rather than mixed environmental samples. Machine-learning targets are genus-level, with the family-only taxon recorded as \textit{Unknown}; expert-verified taxon identifications are retained at the most defensible rank. Although the French, Hungarian, and Swedish crops benefit from extended-depth slide material, the released records are still single 2-D crop views, and the Mediterranean material is not equivalent in this respect. Some aperture or ornamentation details may therefore remain uncertain. The corpus is also geographically biased toward European taxa, and class abundance is highly imbalanced.

The caption and benchmark layers have their own limitations. The captions are anchor-guided machine outputs and may inherit anchor bias or describe attributes that are plausible for a taxon but not visible in a specific crop, a known risk for generative vision--language models. A systematic expert audit of a Gemma4 caption subsample is not included in our work; the captions are therefore best treated as generated morphological benchmark text whose aggregate compliance and retrieval behaviour have been measured, rather than as per-grain expert descriptions. The retrieval results should also be interpreted within the pure-species, expert-anchored benchmark used here: caption prompts and independently rephrased retrieval queries both draw on standard palynological vocabulary, and retrieval evaluates within the assembled atlas rather than on unanchored mixed environmental samples. The results therefore show that adding a morphology-focused text layer is a promising direction for robust retrieval and quality control, not that the present system solves zero-shot operational pollen detection. The VLM comparison is partly family-confounded because four of the five models are Qwen-family models and only Gemma4 provides a cross-family comparator. Retrieval uses a general-purpose SBERT encoder\cite{Reimers2019SentenceBERTSE}, equal fusion uses a naive \(\alpha=0.5\) weight, classification is limited to frozen-feature linear probing, and CROSS-REG retrieval covers 15 genus-level targets. These choices make the benchmarks transparent and reproducible, but they leave clear room for pollen-specific encoders, stronger fusion, and future expert review.

Next steps could use the atlas both as training material and as an evaluation resource. For the visual backbone, a direct route is systematic pollen-domain fine-tuning of ViT-S/B/L-scale dense encoders on mined crops and hard negatives, followed by testing under the same cross-regional benchmark. Such models should first be benchmarked on pure-species reference material, but the same dense-mining principle is not restricted to reference slides: with suitable validation, active learning, and mixed-sample ground truth, it could also support candidate discovery in operational environmental samples. This is more informative for whole-slide deployment than assuming that newer or larger generic encoders will automatically solve the mining problem. For language, the structured captions provide two complementary training signals: free-text supervision for pollen-specific VLMs, projectors, or rerankers that connect visual evidence to standardised morphology vocabulary, and parseable morphology terms that could be converted into attributes such as aperture type, ornamentation, and outline for multi-task image supervision\cite{Zhang2026PollenAttributes}. Raw VLM vision tokens should be evaluated in this setting rather than assumed to replace DINO-style mining features, because VLM vision towers and self-supervised dense encoders are optimised for different objectives. A credible future system could combine DINO-like dense candidate generation with VLM-based semantic quality control and a pollen-adapted vision--language model, in which a pollen-specific visual backbone is connected to an LLM through a trainable alignment module. Such aligned models may outperform image-only baselines, but this should be demonstrated by explicit training and held-out-origin testing, and will likely benefit from extending the pure-species reference-slide design to additional taxa, regions, preparations, and scanner conditions. Such a model could represent apertures, ornamentation, outline, and polarity in a shared multimodal space rather than only describing them after detection\cite{conch_mahmoodlab}. Future versions could broaden geographic coverage, include mixed environmental samples, and use active learning to prioritise uncertain cases for expert review. More broadly, the same assembly pattern--one-shot expert initialisation, dense mining, VLM captioning, and sampled expert validation--should transfer to other microscopy domains where exhaustive manual annotation is unrealistic. Such systems should be framed as expert-support tools, not expert replacements: expert palynologists remain the interpretive reference standard, especially in ambiguous or high-stakes cases. The atlas is therefore best viewed not as the endpoint of automated pollen recognition, but as high-precision infrastructure for future pollen-adapted AI systems and expert-guided multimodal analysis.

\section*{Conclusion}
Pollen AI Atlas demonstrates that expert-guided foundation-model workflows can transform pure-species whole-slide pollen microscopy into a high-precision, million-scale multimodal reference resource. Using one manually selected exemplar per source slide, dense ViT token retrieval, SAM segmentation and multi-stage filtering, the workflow produced 1.5 million released grain detections with a measured false-positive rate below 0.5\%. Expert-verified palynological anchors further enabled controlled morphological captioning at scale, linking each grain to structured descriptions of apertures, ornamentation, shape and size. Downstream benchmarks highlight the complementary value of images and text: frozen visual features provide a strong supervised classification baseline, while morphology-focused text is more robust under cross-regional scanner and slide-preparation shifts. Rather than serving as an exhaustive detector for mixed environmental samples, the atlas is intended as high-precision infrastructure for pollen recognition, validation, domain adaptation and future pollen-adapted vision--language models. More broadly, it demonstrates how limited expert input---focused on exemplar selection, anchor curation and sampled validation---can be combined with automation to create scalable, reusable resources for multimodal microscopy AI.

\section*{Methods}

\begin{figure}[ht!]
\centering
\includegraphics[width=0.999\textwidth]{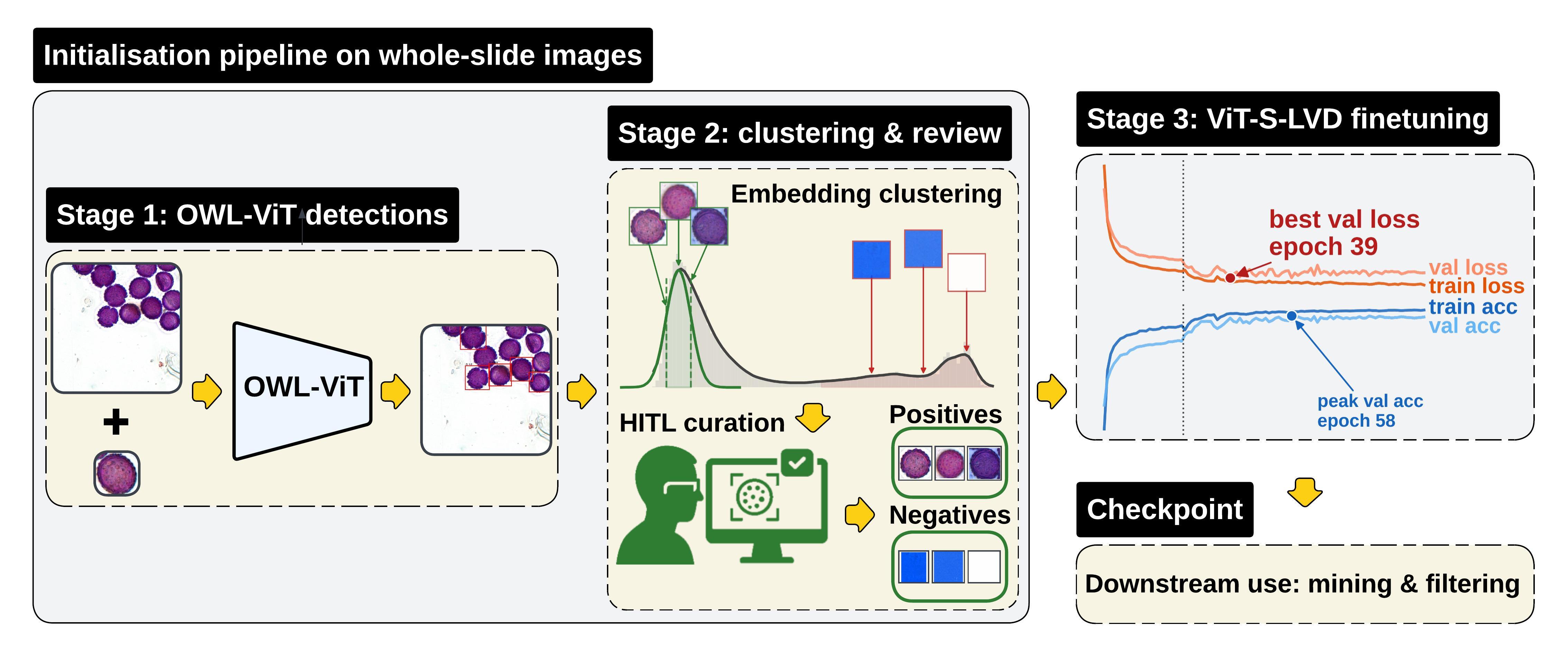}
\caption{One-shot initialisation workflow on whole-slide data. For each slide, Stage~1 uses a manually selected pollen-grain exemplar as a visual query for OWL-ViT one-shot detection, producing candidate crop proposals from the whole-slide image. Stage~2 embeds the proposals, applies similarity-based clustering, and uses human-in-the-loop review to curate positive pollen crops and negative background or artefact examples. Curated crops are then pooled for Stage~3, in which the LVD-ViT-S backbone is fine-tuned and the checkpoint selected by validation loss is retained for downstream token-level mining and quality filtering.}
\label{fig:init_pipeline}
\end{figure}

\begin{figure}[ht!]
\centering
\includegraphics[width=0.943\textwidth]{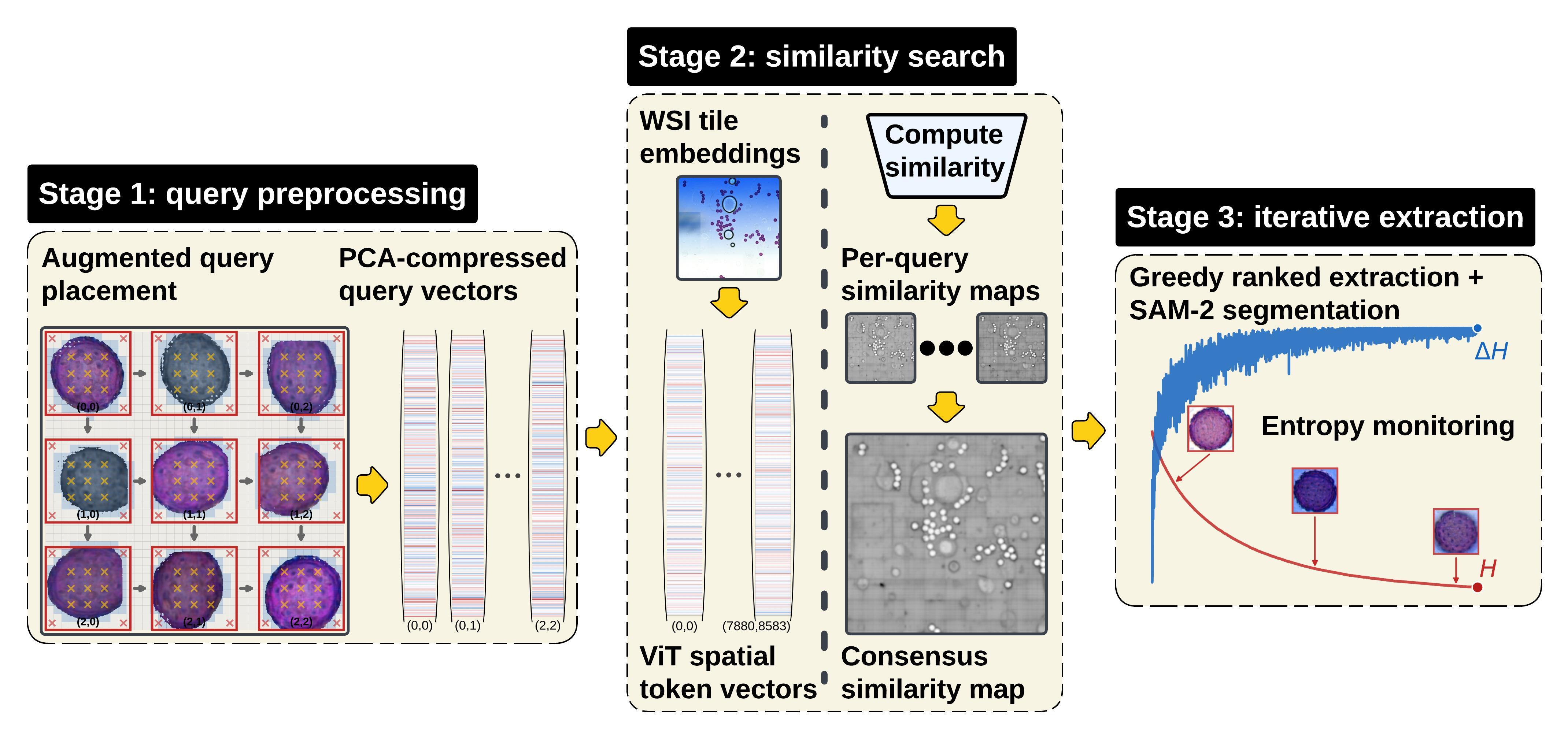}
\caption{Token-level mining workflow for whole-slide pollen discovery. Stage~1 constructs a token-space query representation from a manually selected exemplar grain. The exemplar is placed at multiple shifted positions within the 518$\times$518\,px ViT input field, augmented, segmented with SAM-2 point prompts, and restricted to foreground token vectors, which are then PCA-compressed into a set of query vectors. Stage~2 applies the fine-tuned LVD-ViT-S backbone to compute dense spatial token embeddings across whole-slide tiles. Token-wise cosine similarity between query vectors and whole-slide token embeddings produces per-query similarity maps, which are aggregated into a consensus similarity map. Stage~3 performs greedy ranked extraction of candidate objects from the consensus map, segments each candidate with SAM-2, and monitors convergence using Shannon entropy during iterative mining.}
\label{fig:mining_pipeline}
\end{figure}

\begin{figure}[ht!]
\centering
\includegraphics[width=1.000\textwidth]{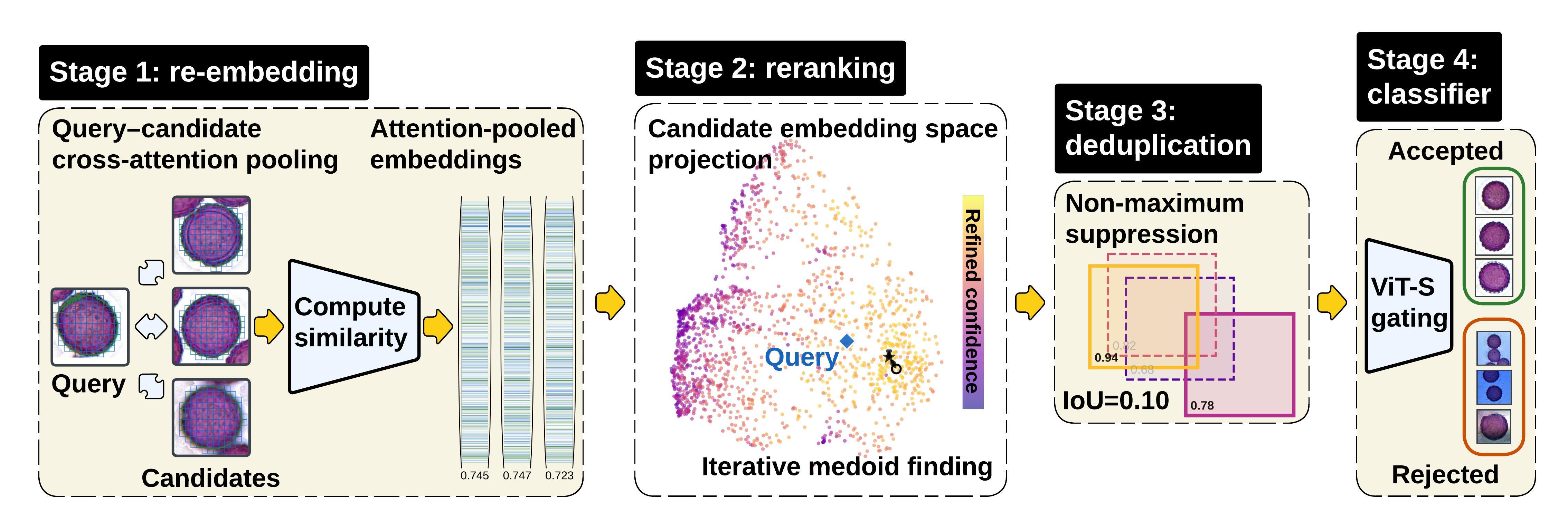}
\caption{Quality-filtering workflow applied to raw mining detections. Stage~1 re-embeds each query--candidate pair by query--candidate cross-attention pooling, producing uniform attention-pooled embeddings for downstream comparison. Stage~2 reranks candidates through iterative medoid-based prototype refinement in embedding space; the two-dimensional plot schematically represents candidate embeddings, with point colour indicating refined confidence after rescoring against the converged prototype. Stage~3 removes spatially overlapping detections by non-maximum suppression using an IoU threshold of 0.10. Stage~4 applies the fine-tuned LVD-ViT-S classifier as a final semantic gate, accepting pollen candidates and rejecting background or low-confidence detections. Accepted detections form the filtered input for downstream caption generation.}
\label{fig:filtering_pipeline}
\end{figure}

Dataset generation treated pure-species whole-slide microscopy as a searchable, expert-guided feature-space resource rather than a manually enumerated object list. The workflow comprised five stages: one-shot initialisation (Fig.~\ref{fig:init_pipeline}), token-level mining (Fig.~\ref{fig:mining_pipeline}), quality filtering (Fig.~\ref{fig:filtering_pipeline}), vision--language captioning (Fig.~\ref{fig:captioning_pipeline}), and expert ground-truth annotation. The procedures used to generate and process the dataset are described below, followed by the construction of the train and validation splits used for downstream evaluation, and the experimental protocols for caption quality assessment, classification, and retrieval. Morphological terminology follows standard palynological usage and the recommended nomenclature for aerobiological monitoring\cite{terminology_paper_pollen}.

\subsection*{Data acquisition}

Whole-slide images were drawn from four independent pollen collections comprising French, Hungarian, Mediterranean, and Swedish material (Supplementary Table~\ref{tab:dataset_summary}). Most slides contained pollen assigned to a single identified taxon, whereas a small number were available only with family-level labels. Together, the collections capture variation in specimen origin, slide preparation, scanner systems, magnification, and effective pixel size. Supplementary Table~\ref{tab:dataset_summary} summarises the collections included in the released dataset. Per-slide pixel calibration values are provided in the deposited calibration file.

\subsubsection*{French collection}

The French collection comprised 49 reference slides derived from curated pollen material. Each slide contained pollen assigned to a single identified taxon. Slides were digitised on a 3DHistech Pannoramic 250 Flash III scanner at 20$\times$ magnification, yielding an effective resolution of 0.242797\,\textmu m\,px$^{-1}$. Ten focal planes spaced at 1\,\textmu m intervals were acquired for each field and fused into extended-depth-of-field (EDF) composites using a custom Dual-Tree Complex Wavelet Transform (DTCWT)-based workflow\cite{dtcwt_article, Ravi2024}.

\subsubsection*{Hungarian collection}

The Hungarian collection comprised five reference slides prepared from single-taxon pollen material. Three slides were digitised on a 3DHistech Pannoramic 250 Flash III scanner at 20$\times$ magnification (0.242797\,\textmu m\,px$^{-1}$), and two slides were digitised on a 3DHistech Pannoramic 1000 scanner at 40$\times$ magnification (0.121295\,\textmu m\,px$^{-1}$). For all Hungarian slides, ten focal planes were acquired and fused into EDF composites using the same DTCWT-based workflow.

\subsubsection*{Mediterranean collection}

The Mediterranean collection comprised 11 slides obtained from a publicly available pollen reference repository\cite{yolo_rcnn_example_Chaves2024_webview}. Each slide was associated with a single dominant identifiable taxon representative of Mediterranean flora. These slides were acquired on an Olympus VS120 scanner at 40$\times$ magnification (NA~0.95) with an effective resolution of 0.17\,\textmu m\,px$^{-1}$. The available whole-slide images consisted of single focal layers and were reassembled from deposited image tiles.

\subsubsection*{Swedish collection}

The Swedish collection comprised 20 reference slides from the Swedish Museum of Natural History. Each slide contained pollen assigned to a single identified taxon. The material includes wind-pollinated tree, grass, and herbaceous taxa representative of Scandinavian aerobiology. Slides were digitised on an Objective Imaging system. After post-acquisition processing, seven slides were retained at effective 40$\times$ resolution (0.1225\,\textmu m\,px$^{-1}$), and thirteen were retained at effective 20$\times$ resolution (0.243\,\textmu m\,px$^{-1}$). Where vendor-generated EDF composites were unavailable, focal planes were fused using the same Dual-Tree Complex Wavelet Transform (DTCWT)-based workflow\cite{dtcwt_article, Ravi2024}.

\subsubsection*{File-format harmonisation}

Scanner outputs in proprietary or BigTIFF formats were converted to pyramidal TIFF using libvips\cite{libvips_github} to enable uniform access through OpenSlide\cite{Goode2013OpenSlide} and TiffSlide\cite{Poehlmann2022TiffSlide}. Native effective pixel sizes were preserved. For the 3DHistech slides, pixel spacing was obtained directly from embedded TIFF metadata. For the Olympus and Objective Imaging collections, per-slide effective resolutions were assigned from scanner specifications or collection metadata and were cross-checked against expected grain-size ranges, as documented in the deposited calibration records.

\subsubsection*{Slides excluded from caption generation}

All raw 85 slides from all collections were processed through mining and filtering, but five were excluded from the caption-generation stage. Four slides carried only nominal family-level labels inherited from the source collections that could not be independently verified in either Beug's \textit{Leitfaden der Pollenbestimmung}\cite{beug2021leitfaden} or the PalDat database\cite{PalDat2000}, even at family level. Taxon-specific caption anchors were therefore not constructed for these slides. One Swedish \textit{Betula} slide was excluded for low image and staining quality. Detection outputs (bounding boxes, masks, and confidence scores) for all five excluded slides are included in the data deposit. One additional family-level slide (Fabaceae) was retained for caption generation with family-level guidance, with its genus-level label recorded as \textit{Unknown}. In total, 80 slides were included in the caption-generation stage.

\subsection*{One-shot initialisation}

One-shot initialisation was used to generate an initial set of candidate pollen crops and a visual backbone for downstream mining and filtering. The procedure followed an image-guided bootstrapping approach similar to that described in our related detector-oriented study\cite{Biricz2025}, and was adapted here to produce a curated crop set for fine-tuning a ViT-Small classifier and its associated embedding space.

\subsubsection*{Stage 1: OWL-ViT proposal generation}

For each slide, a single manually selected pollen crop was used as a visual exemplar. Representative exemplar seed crops are shown in Supplementary Fig.~\ref*{fig:grain_showcase}. This exemplar was supplied to OWL-ViT\cite{owl_objdet_model_1_2022} (\texttt{google/owlvit-base-patch32}), which was applied to overlapping 896$\times$896\,px whole-slide tiles to generate candidate bounding boxes by image-guided detection. A low confidence threshold of 0.01 was used at this stage to favour recall. Candidate boxes were further restricted by coarse size filtering relative to the query crop. As most source slides contained pollen assigned to a single identified taxon, the resulting candidates inherited the slide-level label and were used as a provisional weakly labelled set for subsequent clustering and review.

\subsubsection*{Stage 2: Embedding-based clustering and manual verification}

Following non-maximum suppression of overlapping proposals, candidate crops were extracted from the whole-slide images, padded to a uniform square size, and embedded using the DINOv2-pretrained LVD-ViT-S/14 backbone\cite{ViT_dosovitskiy2021an, Oquab2023DINOv2LR}. Cosine similarity to the exemplar embedding was used to retain likely inliers and reject obvious debris. An acceptance threshold was estimated by fitting a Gaussian model to the main similarity peak; where a stable fit was not obtained, a percentile-based fallback was applied.

Additional geometric and crop-quality filters, including centring, shape regularity, and patch-size constraints, were applied to remove residual artefacts. From each slide, up to 1,000 candidate pollen crops and up to 1,000 negative crops were sampled and reviewed as thumbnail grids in a file-browser interface. A human expert removed residual false positives and retained representative pollen and background examples. The resulting curated set comprised 92,024 images from 68 slides, including 49,880 pollen crops and 42,144 negative crops across 52 classes based on slide-level labels, and was used for subsequent model training.

\subsubsection*{Stage 3: ViT-Small backbone fine-tuning}

The curated crop set was used to fine-tune the DINOv2-pretrained LVD-ViT-S/14 encoder as a 52-class classifier. The 52-class label space comprised provisional per-slide labels inferred directly from source-slide identifiers during early bootstrapping; these labels predated the subsequent human-in-the-loop taxonomic verification, caption-anchor curation, and exclusion of unresolved records that produced the final 46-label space, comprising 45 genera and \textit{Unknown} for the family-only Fabaceae slide. The classification head used during adaptation was retained for quality filtering, where the same checkpoint was reused solely as a pollen-versus-background semantic gate rather than as a fine-grained taxonomic classifier. Training and validation subsets were defined using a stratified 80:20 split. Optimisation was performed in two phases. During epochs 0--23, the backbone was frozen and only the linear classification head was trained. During epochs 24--99, all parameters were jointly optimised using differential learning rates of $5\times10^{-6}$ for the backbone and $1\times10^{-4}$ for the classification head. AdamW optimisation\cite{AdamW} and CosineAnnealingLR scheduling were used throughout. Training was run for 100 epochs, and the checkpoint with the lowest validation loss was selected for mining and filtering. The pollen-adapted checkpoint was used for token-level mining and quality filtering, whereas downstream classification and image retrieval used frozen features from the original LVD-ViT-S/14 backbone.

\subsection*{Token-level mining}

Token-level mining was used to generate candidate pollen detections across the whole-slide images. In total, 1,826,529 candidate regions were produced from 85 slides. This procedure treated the spatial token grid of the classification-finetuned LVD-ViT-S/14 backbone as a searchable feature map. A single exemplar crop was used to construct a query representation in token space, which was matched against all whole-slide tokens using cosine similarity. High-scoring locations were converted into object candidates by iterative SAM-2 segmentation. The resulting candidates were used as input to the subsequent filtering stage.

\subsubsection*{Stage 1: query preprocessing}

For each slide, a single exemplar crop selected during the initialisation stage was used to construct the token-space query for mining. Supplementary Fig.~\ref*{fig:grain_showcase} illustrates representative seed exemplars for this mining stage. The crop was first segmented with SAM-2\cite{Ravi2024SAM2S} to isolate the pollen foreground. In the standard production setting, query segmentation used a multi-point prompt comprising nine positive points arranged on a 3$\times$3 internal grid and four negative points placed near the image corners. Only ViT tokens whose receptive fields overlapped the resulting foreground mask were retained as query tokens.

Query preprocessing was designed to reduce the dependence of token activations on the spatial placement of the pollen grain within the ViT input field. The fine-tuned LVD-ViT-S/14 backbone accepts 518$\times$518\,px inputs and partitions each input into a fixed 37$\times$37 grid of 14$\times$14\,px patches. Because most exemplar grains occupied only a limited part of this field, otherwise similar grains could intersect patch boundaries differently when appearing at different positions within a whole-slide tile, leading to position-dependent token patterns. To reduce this effect, the exemplar crop was pasted onto a 518$\times$518\,px canvas at multiple regularly spaced positions and re-encoded at each placement. In the standard production configuration, this placement grid comprised 10$\times$10 positions, yielding 100 shifted query views; some earlier runs used 8$\times$8 placements, and the exact settings used for each slide are recorded in the deposited mining statistics.

Each shifted placement was additionally subjected to appearance augmentation before re-encoding. The augmentation scheme included right-angle rotation, horizontal and vertical mirroring, colour jitter, greyscale conversion, affine perturbation, and Gaussian blur. The first shifted query view was kept unaugmented in order to preserve the canonical exemplar representation. After placement and augmentation, each query view was re-segmented with SAM-2, and only foreground tokens were retained for subsequent aggregation.

For each shifted query view, the retained foreground tokens were reduced to a single 384-dimensional query vector by top-$k$ principal-component aggregation, with $k=10$ in production. This aggregation was performed by computing the principal components of the masked token set, aligning component signs relative to the token mean, and combining the retained components with weights proportional to explained variance. The resulting vector therefore summarised the dominant directions of variation present in the masked exemplar tokens for that placement. Each query vector was then used independently in the next stage.

\subsubsection*{Stage 2: whole-slide token database and similarity-map construction}

Each whole-slide image was converted into a slide-wide token database by partitioning it into non-overlapping 518$\times$518\,px tiles and passing each tile once through the fine-tuned LVD-ViT-S/14 backbone. For every tile, the backbone produced a 37$\times$37 grid of spatial tokens, corresponding to 1,369 token embeddings of 384 dimensions. Across a typical slide, this yielded approximately 30{,}000--58{,}000 tiles, corresponding to about 45--80 million token embeddings, depending on image size. The token embeddings were stored together with their spatial coordinates so that each token could be mapped back to its location in the whole-slide image during subsequent ranking and extraction.

Each 384-dimensional query vector generated during Stage~1 was then matched independently against this slide-wide token database. Cosine similarity was computed between the query vector and all whole-slide token embeddings. Because each slide contained tens of millions of tokens, similarity computation was performed on the GPU in chunked batches. For each query vector, the resulting similarity scores were scattered back onto the spatial token grid of the whole-slide image to form a slide-wide similarity map.

Before aggregation, each similarity map was normalised to the unit interval, smoothed with a Gaussian filter, and renormalised so that it could be interpreted consistently across query views. The multiple similarity maps obtained from the shifted and augmented query representations were then aggregated by location-wise median to form a single consensus similarity map. Median aggregation was used to reduce the influence of individual query views that failed because of unfavourable spatial placement, weak augmentation, or imperfect query segmentation, while retaining locations that were supported consistently across the set of query representations.

The final consensus map was renormalised and treated as a discrete probability distribution over token locations. This representation was used in the next stage to rank candidate locations for extraction and to monitor convergence of the iterative mining process by Shannon entropy.

\subsubsection*{Stage 3: iterative mining loop}

Mining proceeded greedily over the ranked consensus similarity map. Token locations were traversed in descending order of consensus similarity, subject to a slide-specific upper-tail percentile cutoff recorded in the deposited mining statistics. At each iteration, the highest-scoring token that had not yet been marked as visited was selected as the next candidate location. If a token had already been visited, the iteration was skipped without repeating extraction or segmentation.

A square image region centred on the selected token was then extracted on demand from the whole-slide image. The extraction size was scaled relative to the exemplar crop by setting the mining tile size to 2.5 times the maximum exemplar dimension. This centred, on-demand extraction avoided the use of a fixed precomputed detection grid and reduced boundary artefacts by keeping candidate grains close to the middle of the extracted region.

SAM-2 was then applied at the selected location to generate a candidate object mask within the extracted region. Depending on the production run, segmentation was performed either in single-mask mode or in multimask mode. In multimask mode, SAM-2 returned multiple candidate masks, and the final mask was chosen using the model score together with size constraints defined relative to the query object. Both the mask pixel area and its bounding-box area were required to remain within query-relative bounds. Small disconnected components and small holes were removed during mask post-processing before further evaluation.

For accepted candidate masks, the corresponding whole-slide token embeddings were gathered and averaged to form an object-level embedding. This embedding was compared by cosine similarity to a running cluster mean to obtain a candidate-level confidence score. In runs where confidence screening was enabled, low-confidence candidates were discarded before storage. To stabilise this reference during mining without retaining all previously accepted objects in memory, the running cluster mean was updated from a bounded reservoir containing the highest-confidence object embeddings observed so far. This prototype was frozen once successive updates became negligible, or once the reservoir reached its preset capacity.

Similarity-map updates were applied sparsely. For accepted masks, all covered token locations were zeroed in the consensus map so that they could not be selected again. For rejected masks, only the queried token location was set to the similarity floor. To avoid the cost of repeated full-map deduplication, non-maximum suppression was applied periodically in GPU batches rather than at every iteration, and the proportions of accepted, replaced, and discarded candidates were tracked throughout the run.

\subsubsection*{Stage 4: convergence monitoring and stopping criteria}

Mining progress was monitored by Shannon entropy over the normalised consensus similarity map,
\begin{equation}
H = -\sum_i p_i \ln p_i,
\end{equation}
where $p_i$ denotes the normalised similarity mass at token location $i$. At the start of mining, similarity mass is distributed across many token locations and entropy is correspondingly high. As candidate objects are accepted and their covered tokens are removed from the map, the remaining distribution becomes progressively less informative and entropy generally decreases.

Because full recomputation of entropy over tens of millions of token locations at every iteration would be computationally prohibitive, entropy was updated incrementally from only the token values modified at that step. This reduced the update cost from a full-map operation to a sparse local update involving only the small subset of tokens affected by mask acceptance or rejection. Windowed entropy differences were then computed over the mining trajectory and smoothed by exponential moving averaging to provide a stable estimate of entropy slope.

Termination was governed primarily by sustained flattening of the entropy decrease. When successive iterations removed little additional entropy over a prolonged interval, the mining run was considered to have reached convergence and was stopped. Auxiliary stopping criteria were also used to terminate unproductive runs. These included exhaustion of the ranked candidate list above the configured percentile threshold, persistently high duplicate-discard ratios during periodic non-maximum suppression, and persistently high bad-mask rejection ratios during segmentation. Across all 85 slides, token-level mining required around 500 GPU-hours on A100 hardware. Density-sensitive parameters, including the percentile cutoff, entropy-slope multiplier, mask-ratio bounds, and related stopping settings, were tuned across production runs and are reported per slide in the deposited mining statistics.

\subsection*{Quality filtering}

A four-stage filtering pipeline was applied to the raw mining output to remove residual debris, fused clusters, spatial duplicates, and low-confidence detections. Across all 85 slides, this procedure reduced 1,826,529 raw mining candidates to 1,528,984 filtered detections. The four stages were applied in fixed order: attention-based re-embedding, prototype-based reranking, non-maximum suppression, and classifier gating. The filtered detections formed the direct input to the captioning workflow. After subsequent slide-level taxonomic review, 1,511,390 detections from 80 slides were taken forward to caption generation, whereas five filtered slides were excluded from captioning as described above.

\subsubsection*{Stage 1: attention-based re-embedding}

The confidence values produced during token-level mining were not directly comparable across the full discovery sequence, because candidate objects were encountered at different stages of an iterative search process. To construct a uniform per-candidate representation for subsequent reranking, each candidate was re-embedded by attention pooling between the canonical query tokens and the token embeddings stored for that candidate during mining. For each candidate, the ViT tokens overlapping its segmentation mask were gathered directly from the whole-slide embedding database; when a candidate's token set exceeded 1,024 elements, it was reduced to the 1,024 tokens with the highest similarity to any query token in order to bound memory use. The retained tokens were pooled by scaled dot-product attention independently for each query token. The resulting pooled representations were averaged across query tokens and normalised to unit length, yielding a single attention-pooled embedding per candidate.

\subsubsection*{Stage 2: prototype-based reranking}

The hand-selected exemplar used during mining was not assumed to lie at the centre of the morphological distribution represented by the candidate set. To reduce sensitivity to exemplar-specific bias, the attention-pooled embeddings from Stage~1 were used to estimate a refined slide-level prototype by iterative medoid computation. At each iteration, all candidates were scored against the current prototype, the top 50\% were retained, and the medoid of that subset was recomputed in embedding space. Iteration continued until the prototype change fell below a convergence tolerance of $10^{-4}$ or the maximum number of iterations was reached. For large subsets, medoid computation was performed in chunks in order to avoid construction of a full pairwise similarity matrix. After convergence, all candidates were rescored by cosine similarity against the final prototype, and these refined confidence values were used in all subsequent filtering stages.

\subsubsection*{Stage 3: non-maximum suppression}

Spatially overlapping detections were then removed by non-maximum suppression using the refined confidence values from Stage~2 and an IoU threshold of 0.10. This threshold was chosen to retain a single representative crop per physical grain while suppressing duplicates arising from nearby similarity peaks or repeated segmentation of the same object. Because reranking had already established a slide-wide confidence ordering, the highest-quality candidate among overlapping detections was preferentially retained. A separate post-NMS confidence threshold was not applied in production.

\subsubsection*{Stage 4: classifier gating}

The remaining candidates were evaluated with the fine-tuned LVD-ViT-S/14 classifier obtained during the initialisation stage. For each object, a context-padded square crop was extracted from the whole-slide image using a padding factor of 1.2 relative to the bounding-box extent, resized to 518$\times$518\,px, and scored with the provisional 52-category bootstrapping head described above. Whole-slide crop extraction was parallelised across CPU worker threads, and classification was performed in GPU batches.

Candidates predicted as background were removed. Candidates assigned to a pollen class were retained only if the predicted-class probability was at least 0.5. This final stage served as the principal semantic gate against residual debris, partial objects, fused clusters, and other non-pollen detections that survived the preceding geometric and embedding-based filters.

\subsubsection*{Filtered outputs}

Filtered detections were stored as per-slide HDF5 files. For each retained object, the output included the segmentation mask, prompt point, crop origin, bounding box, original mining confidence, refined confidence, classifier prediction, predicted-class probability, predicted class name, and rebuilt token-index metadata inherited from the mining stage. Run-level mining trajectories, including entropy and confidence traces, were preserved alongside the retained objects. These filtered HDF5 files formed the direct input both to downstream caption generation and to the proposal-based expert annotation workflow.

\subsection*{VLM captioning}

\begin{figure}[ht]
\centering
\includegraphics[width=0.917\textwidth]{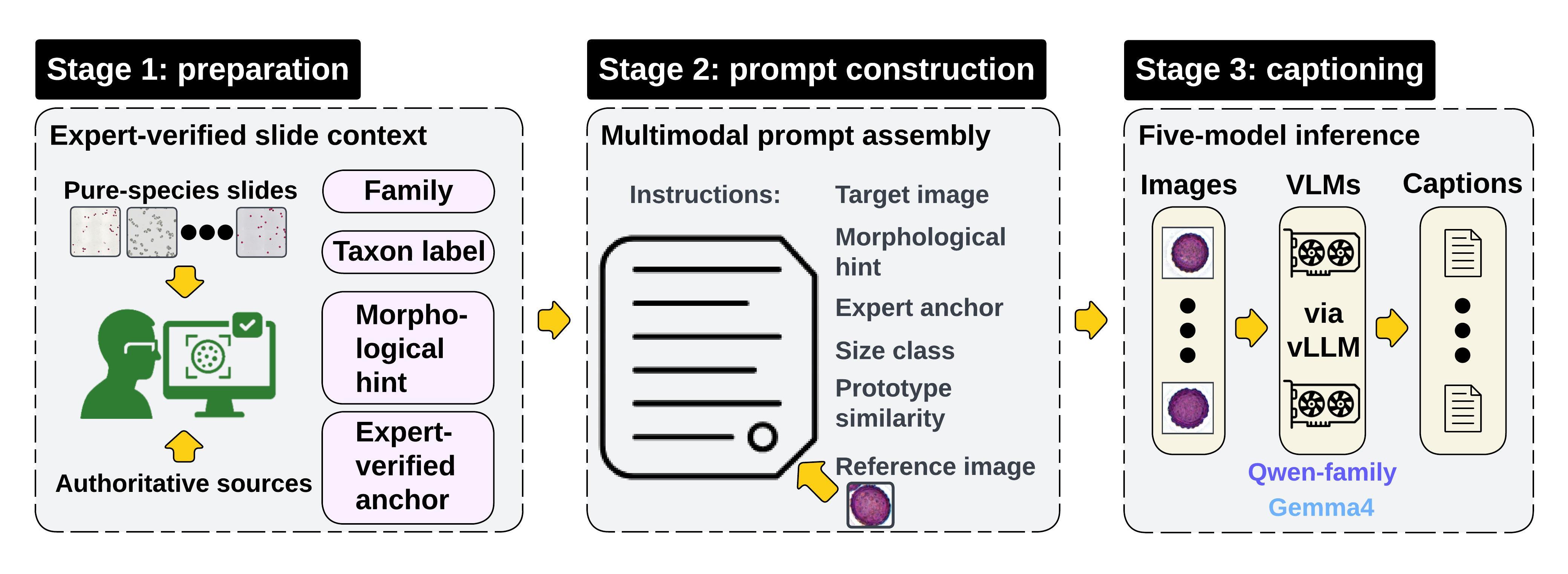}
\caption{VLM captioning workflow. Stage~1 prepares slide-specific captioning context from pure-species slides and authoritative palynological sources, yielding family and taxon assignments, a condensed morphological hint, and an expert-verified anchor description derived primarily from Beug's \textit{Leitfaden der Pollenbestimmung} with PalDat as a supplementary source. Stage~2 assembles the multimodal prompt by combining detailed text instructions with the target grain image, an exemplar reference image, the morphological hint, the anchor description, a qualitative size class, and a prototype-similarity confidence value. Stage~3 performs independent caption generation with five open-weight vision--language models, comprising four Qwen-family models and Gemma4, served through vLLM. Each retained detection received one caption per model, producing five grain-level caption sets for downstream diagnostics and multimodal benchmarking.}
\label{fig:captioning_pipeline}
\end{figure}

The captioning stage added a text modality to the filtered pollen detections by generating structured morphological descriptions for each retained grain. Each of the 1,511,390 detections taken forward to caption generation was processed independently by five open-weight vision--language models using the same slide-specific multimodal prompting protocol (Fig.~\ref{fig:captioning_pipeline}). The protocol combined the target grain crop, a slide-level exemplar image, and expert-verified palynological anchor material to produce one caption per model for each detection, yielding 7,556,950 captions across 80 slides.

\subsubsection*{Stage 1: preparation}

For each of the 80 captioned slides, four manually curated anchor files were prepared: a genus-level target label (or \textit{Unknown}), a botanical family, a short morphological hint, and a longer anchor description. The anchor descriptions were derived primarily from Beug's \textit{Leitfaden der Pollenbestimmung}\cite{beug2021leitfaden}, the standard central European pollen identification guide, with the PalDat database\cite{PalDat2000} as supplementary source for taxa not covered by Beug. All anchors were verified through a human-in-the-loop review process: the domain expert (B.G.) reviewed and verified each anchor for taxonomic correctness, and the first author cross-checked all entries against the same reference. The anchor texts were further cross-checked against \textit{Illustrated Pollen Terminology}\cite{Pollen_book_terminology} and the corresponding whole-slide imagery. The anchor system provided standardised palynological vocabulary and slide-specific reference context for prompt construction, so that caption generation remained tied to established terminology rather than unconstrained free-text description.

\subsubsection*{Stage 2: prompt construction}

Each captioning request comprised two images: the target grain crop, extracted at native resolution from the whole-slide image using the filtered bounding box with a small margin, and an exemplar reference image representing the same slide-level taxon. The text prompt instructed the model to produce a single-paragraph morphological description of 60--80 words covering aperture system, wall ornamentation, shape, and qualitative size class. Taxon names were explicitly prohibited in the output to prevent label leakage, and the prompt further instructed the model not to report raw numeric measurements verbatim.

In addition to the two images, the prompt included four contextual inputs: (i)~a short morphological hint summarising key diagnostic features; (ii)~a longer reference description derived primarily from Beug's \textit{Leitfaden der Pollenbestimmung}\cite{beug2021leitfaden}, with the PalDat database\cite{PalDat2000} as supplementary source for taxa not covered by Beug, providing standardised palynological vocabulary and cross-checked by the domain expert against \textit{Illustrated Pollen Terminology}\cite{Pollen_book_terminology}; (iii)~a qualitative size class derived from segmentation-mask measurements and per-slide pixel-to-micron calibration; and (iv)~a prototype-similarity confidence value supplied as a typicality prior. Quantitative measurements, including polar diameter, equatorial diameter, and grain area converted to micrometre units from the segmentation mask, were provided in the prompt context but were accompanied by instructions not to quote them directly in the caption text. An explicit debris-opening sentence was specified for objects clearly inconsistent with pollen morphology. Production inference used a sampling temperature of 0.3 and a maximum output length of 250 tokens.

\subsubsection*{Stage 3: captioning}

Five open-weight VLMs were used for production inference. Four are successive releases from the Qwen-VL family: Qwen2.5-VL-32B-AWQ\cite{bai2025qwen25vltechnicalreport} (32B parameters, 4-bit AWQ quantisation, ${\sim}$17\,GB VRAM per GPU), Qwen3-VL-32B-Instruct-FP8\cite{bai2025qwen3vltechnicalreport} (32B parameters, a dense Qwen3-VL checkpoint with FP8 quantisation, ${\sim}$31\,GB VRAM per GPU), Qwen3.5-27B-FP8\cite{QWEN35_27B} (27B parameters, unified early-fusion VLM, FP8 quantisation, ${\sim}$27\,GB VRAM per GPU), and Qwen3.6-27B-FP8\cite{QWEN36_27B} (27B parameters, FP8 quantisation, ${\sim}$27\,GB VRAM per GPU). The fifth model, Gemma-4-31B-it\cite{gemma4_31b_it} (31B parameters, served in BF16 precision, approximately 60 GB VRAM per GPU), was included as a cross-family comparator from a distinct architecture and training regime. The four Qwen generations enable within-family generational comparison, while the Gemma model provides a between-family comparison. Gemma4 was selected as the main VLM for downstream experiments based on low taxon and numeric leakage, high prompt compliance, broad anchor-vocabulary coverage, and the strongest ALL-mode text-retrieval performance.

Caption generation was performed with vLLM\cite{Kwon2023EfficientMM} using A100-SXM4-80\,GB GPUs on the compute cluster and RTX~4090 GPUs on the local workstation. On the cluster, each A100 GPU hosted a separate vLLM inference server with tensor-parallel size one. An asynchronous Python client distributed requests across available servers with bounded concurrency, round-robin server rotation, and exponential-backoff retry logic after failed requests. Captioning was processed in batches for memory management, and intermediate outputs were checkpointed periodically to support recovery and resume.

The complete five-model production run across 80 slides generated 7,556,950 captions. Across the five models, production captioning required approximately 125 wall-clock hours on the 8$\times$A100-SXM4-80\,GB cluster, corresponding to approximately 1,000 GPU-hours. Captions were written as per-slide JSONL files, with one record per detection and model. Each record included the grain identifier, taxonomic fields, source slide, spatial metadata, quality scores, morphometric fields, and raw model-generated caption text.

\subsection*{Expert ground-truth annotation}

An expert-curated ground truth was constructed to assess detection quality and to provide test labels for downstream classification and retrieval experiments.

\subsubsection*{Test-region selection}

One representative 5,000$\times$5,000\,px region was selected on each of 84 annotated slides. Region selection was based on the spatial distribution of filtered pipeline detections rather than on exhaustive full-slide manual inspection. Candidate regions were ranked by local detection density, and a region close to the 60th percentile of that density distribution was selected in order to avoid both unusually sparse and unusually crowded extremes. The selected region, together with all pipeline detections falling within it, was exported as a QuPath-compatible GeoJSON proposal set.

\subsubsection*{Annotation workflow}

A proposal-correction workflow was used to reduce expert effort. For each selected region, the corresponding whole-slide image and the exported GeoJSON proposals were loaded into QuPath\cite{Bankhead2017QuPath}. The domain expert reviewed each proposed object, retaining correct detections, deleting false positives, and adding missed pollen grains as rectangular annotations. Added objects were classified as pollen by expert inspection and subsequently linked to the expert-verified slide-level taxon metadata. This procedure produced a curated reference set whose primary quantitative subset spanned 45 genera and one family-only taxon across 31 botanical families.

\subsubsection*{Curation and reconciliation}

The corrected QuPath exports were processed by a reconciliation script that compared the expert-edited annotations with the original proposal set through bounding-box matching. Objects preserved by the expert were recorded as confirmed detections, deleted proposals as false positives, and expert-added objects as false negatives. The curated test-region boundary was preserved in the reconciled GeoJSON output.

\subsubsection*{Ground-truth scale}

The complete annotated set, referred to as Test Set 2 (TS2), comprises 10,595 pollen annotations across 84 slides. Of these, 10,335 grains across 80 slides form the primary quantitative subset; the remaining four slides carry nominal family-level labels inherited from the source collections that could not be independently verified and are retained for reference only. A legacy evaluation set Test Set 1 (TS1) from our previous study\cite{Biricz2025} is also retained for compatibility and is described in Appendix~B. The proposal-correction workflow required approximately 16 hours of expert annotation time. Curated annotations are released as per-slide GeoJSON files.

\subsection*{Train/test split construction}

The primary test set was defined for evaluation, and the remaining captioned detections were partitioned into training and validation subsets.

\subsubsection*{Primary test set}

The primary expert-curated test set described above comprises 84 annotated slides in total, of which 80 are used for quantitative evaluation: 79 slides with expert-verified taxon identifications at genus level or finer and one slide identified only to the family Fabaceae, for which the genus-level target was recorded as \textit{Unknown}. This evaluated subset contains 10,335 grains across 46 dataset labels. The remaining four slides with nominal, unverified family-level source labels are deposited for reference only.

\subsubsection*{Training and validation splits}

All 1,511,390 captioned detections were loaded, and any sample whose bounding box intersected an expert-defined test region (TS1 or TS2) was excluded from model development. After excluding 12,389 such detections, 1,499,001 samples remained and were partitioned into training (85\%) and validation (15\%) subsets using a fixed random seed of 42, yielding 1,274,118 training samples and 224,883 validation samples. Splits were defined by sample identifier and are model-agnostic, such that the same partition applies to all VLM caption sets. Split files and verification scripts are included in the data deposit.

\subsection*{Caption quality assessment}

Caption quality was assessed with automated diagnostics designed to evaluate prompt compliance, leakage control, terminology use, and operational search utility for each of the five VLM caption sets. The purpose of these diagnostics was not to certify every generated caption as an expert-written morphological annotation, but to test whether the captioning protocol produced controlled, domain-terminology-rich text suitable for downstream analysis.

Prompt compliance was quantified by word-count distributions, the proportion of captions within the requested 60--80 word range, digit-containing outputs, numeric size-measurement leakage, taxon-name leakage, and occurrence of the explicit debris-screening opener. Taxon-name leakage was evaluated against a corpus-wide token set derived from the slide-level taxon and family labels; \textit{Unknown} was excluded from this token vocabulary. Numeric size leakage was counted when captions contained explicit number--unit expressions rather than qualitative size classes.

Morphological vocabulary use was evaluated against a 422-term anchor-derived palynological lexicon covering aperture systems, wall ornamentation, shape, polarity, and size terminology. For each VLM, vocabulary coverage was reported at corpus level as the proportion of lexicon terms observed at least once in the generated caption set. These diagnostics were interpreted as measures of controlled terminology use rather than evidence of per-grain expert correctness.

To assess operational utility, independently rephrased expert descriptors were embedded with SBERT (\texttt{all-MiniLM-L6-v2}) and queried against each model-specific 1,480,653-caption genus-resolved embedding index in ALL mode without origin exclusion. One descriptor was used as a text query for each of 15 genus-level targets. Query descriptors were written separately from the prompt anchors to avoid direct reuse of text supplied during caption generation. Retrieval was evaluated using P@1, mAP@20, and MRR, with equal weight given to each target. The 95\% confidence intervals were estimated from 10,000 target-level bootstrap resamples using the percentile method. These retrieval-based diagnostics were used to compare the practical utility of the five caption sets and to select the primary caption set for downstream multimodal analyses.

\subsection*{Classification experiments}

Classification experiments were used as a supervised utility baseline for genus-level pollen recognition and to test whether Gemma4 caption embeddings provided information beyond image features under fixed model capacity. All experiments used the same train, validation, and test partitions, the same frozen image backbone, and the same class-balancing strategy; no image-backbone fine-tuning was performed during these downstream evaluations.

Three training settings were compared. In the image-only setting, 384-dimensional CLS embeddings were extracted from the original frozen LVD-ViT-S/14 with its weights frozen and passed to a single linear classification head over 45 genus-level target labels. In the privileged-information setting, the 384-dimensional image embedding was concatenated during training with a 384-dimensional SBERT embedding of the corresponding Gemma4 caption, giving a 768-dimensional input. Caption dropout of 30\% was applied during training, and the caption channel was set to zero at inference so that predictions remained image-only. In the distillation setting, an image-only student was trained from the privileged-information teacher using a combined cross-entropy and Kullback--Leibler divergence objective with temperature $T{=}3$ and mixing weight $\alpha{=}0.5$. The distilled student therefore used only image features at inference.

Each setting was evaluated in five geographic configurations: one model trained on all origins jointly and one model trained on each geographic origin separately. Image-only models were additionally evaluated with Macenko stain normalisation as a simple colour-normalisation baseline. Each experiment was repeated with five random seeds (41--45). Class imbalance was handled by square-root-weighted sampling and loss weighting. Because square-root-balanced sampling produced different epoch lengths for the five training scopes, the schedules used 67, 20, 18, 750, and 298 epochs for combined-origin, French-only, Hungarian-only, Swedish-only, and Mediterranean-only training, respectively. Training used AdamW optimisation with cosine annealing. Although the primary test set contains 10,335 expert-curated grains across 80 evaluated slides and 46 dataset labels (45 genus-level targets plus \textit{Unknown} for the family-only Fabaceae slide), records from that slide were excluded from classification training and evaluation, leaving 45 genus-level target labels across 79 slides and 10,218 valid evaluation samples. Specifically, this Fabaceae slide (30,737 filtered detections) contributed to the captioned release corpus but was excluded from classification and retrieval analyses requiring genus-level target labels; this accounts for the 79-slide retrieval index (1,480,653 grains) relative to the 80-slide, 1,511,390-grain release corpus. Results are reported as Top-1 accuracy and macro F1, with mean $\pm$ standard error across seeds.

\subsection*{Retrieval experiments}

Retrieval experiments evaluated semantic nearest-neighbour search in image and caption embedding spaces. This analysis complemented supervised classification by testing whether query embeddings could recover taxonomically matching grains from the released corpus, including under leave-origin-out conditions that probe domain shift across scanner, preparation, staining, and source-collection differences.

Three query modalities were evaluated. Image queries used exemplar grain crops embedded with the original frozen LVD-ViT-S/14 backbone, yielding 384-dimensional L2-normalised vectors. Text queries used independently written morphological descriptors embedded with SBERT (\texttt{all-MiniLM-L6-v2}), also yielding 384-dimensional L2-normalised vectors. These descriptors were rephrased separately from the caption anchors used during VLM prompting, reducing circularity between prompt text and retrieval query. Combined queries used score-level late fusion of image and text similarities with equal weighting ($\alpha{=}0.5$).

The main retrieval analysis used the Gemma4 caption set, selected from the caption diagnostics as the primary VLM output. The retrieval index contained 1,480,653 grains from the genus-resolved 79 slides with matched image and caption embeddings. Supplementary analyses compared text-only retrieval behaviour across all five VLM caption sets.

Two exclusion modes were defined. ALL used the full index and served as a ceiling condition in which same-slide and same-origin visual similarity could contribute to retrieval. CROSS-REG excluded all grains from the query's geographic origin and served as the primary leave-origin-out benchmark across scanner, preparation, staining and source-collection shifts. Evaluation was restricted to 15 genus-level targets represented in at least two geographic origins. ALL used 44 image and late-fusion queries and 15 text queries. CROSS-REG retained one image query per physical preparation, yielding 40 image and late-fusion queries and 37 genus--origin text evaluations. Retrieval quality was measured by precision at rank 1 (P@1), precision at rank 20 (P@20), mean average precision at 20 (mAP@20), and mean reciprocal rank (MRR). Query scores were first averaged within each genus--origin group, then across origins with equal weight within each genus, and finally across the 15 genera with equal weight. For the unshuffled results, 95\% percentile intervals were estimated from 10,000 hierarchical bootstrap replicates that resampled genera, origins within sampled genera, and queries within sampled genus--origin groups, with replacement at every level. Modality-specific label-shuffle controls were computed across three random seeds as descriptive sanity checks.

\subsection*{Use of AI tools in manuscript preparation}

Large language models from OpenAI (GPT-4o, o3, GPT-5.x family, and Codex) and Anthropic (Claude Opus, Sonnet, and Haiku families) were used during manuscript preparation over the course of the project for drafting assistance, literature search, editing, and code-generation support. All AI-assisted material was reviewed, verified, and edited by the authors. The vision--language models used for grain caption generation are described above as study instruments rather than writing aids. Any AI-assisted drafting, code, or literature-search output was reviewed and verified by the authors; all statistical computations and result interpretation were checked by the authors.

All schematic figures were created by the authors using Python-generated visualisations, real microscope crops from the study data, and the Lucid diagramming application.

\section*{Data availability}

The processed data records described in this paper are deposited in Zenodo at \footnote{\url{https://zenodo.org/records/20690944}}. The public Zenodo record\cite{zenodo_deposit} contains the release archive \texttt{pollen\_ai\_atlas.zip} (42.8\,GB), with file access managed through Zenodo. The package includes filtered detection metadata in HDF5 format; all five caption sets in JSONL format (Qwen2.5-VL, Qwen3-VL, Qwen3.5, Qwen3.6, and Gemma4); expert-curated ground-truth annotations in GeoJSON format; pre-extracted pollen grain crops; validated ground-truth pollen crops; train/validation split definitions; deposited support metadata; caption anchors; query images; and the fine-tuned LVD-ViT-S checkpoint. These records are distributed under a CC BY 4.0 licence. The 85 original processed whole-slide microscopy images (148.9 GB) are not included in the Zenodo package and are available from the corresponding author on reasonable request.

\section*{Code availability}

All custom code used for dataset generation, processing, and technical validation is available at \footnote{\url{https://github.com/abiricz/pollen-ai-atlas-paper}} under an Apache 2.0 licence. The repository contains the one-shot initialisation workflow, token-level mining pipeline, quality-filtering scripts, caption-generation scripts, split-construction utilities, and the evaluation workflows for detection quality, caption quality, classification, retrieval, and embedding visualisation.

A citable versioned archive of the repository will be deposited at time of publication.

The principal software versions used in the current release environment include Python 3.12.3, NumPy 2.2.6, PyTorch 2.10.0, torchvision 0.25.0, timm 1.0.27, Pillow 11.0.0, transformers 5.6.2, vLLM 0.19.1, openslide-python 1.4.1 linked against OpenSlide 3.4.1, TiffSlide 2.5.1, sentence-transformers 5.2.2, scikit-learn 1.5.2, and UMAP-learn 0.5.7. Classification experiments were run with five random seeds (41--45), and the released train/validation partitions were generated with a fixed seed of 42. The released split manifest records the train/validation ratios, sample-ID format, split seed, and split counts; the evaluation configuration and result-generation scripts record the downstream experiment seeds. The fine-tuned LVD-ViT-S checkpoint is included in both the data release and the code release.

\section*{Acknowledgements}

We gratefully acknowledge the contributions of the National Public Health Center (Hungary), the Swedish Museum of Natural History, and the Réseau National de Surveillance Aérobiologique (France) for providing reference materials and supporting data collection.

We extend special thanks to János Fillinger and his team for granting access to their pathology imaging facility. Their expertise in high-resolution slide scanning provided valuable technical input that strengthened this study.

We also thank Viktor Varga for valuable input during the final refinement of the manuscript.

We gratefully acknowledge the Wigner Scientific Computing Laboratory (WSCLAB) for providing the computational infrastructure that made the large-scale model evaluations possible.

A.B. acknowledges carrying out substantial parts of this work without dedicated external funding and with self-funded local computational resources.

\section*{Funding declaration}

This work was supported by the National Research, Development, and Innovation Office of Hungary via grant NKKP-153428 HIGHLIGHT (I.C.) and by the Data-Driven Health Division of the National Laboratory for Health Security (RRF-2.3.1-21-2022-00006) (P.P.).

\section*{Author contributions}

All authors read and approved the final version of the manuscript.\\
\textbf{A.B.}: conceptualisation, data curation, formal analysis, investigation, methodology, project administration, software, validation, visualisation, writing -- original draft.\\
\textbf{B.G.}: project administration (supporting), resources, validation, writing -- review \& editing.\\
\textbf{D.M.}: project administration (supporting), resources, writing -- review \& editing.\\
\textbf{A.S.}: resources, writing -- review \& editing.\\
\textbf{J.F.}: data curation, project administration (supporting), resources, writing -- review \& editing.\\
\textbf{P.P.}: funding acquisition, supervision, project administration (supporting), writing -- review \& editing.\\
\textbf{I.C.}: funding acquisition, supervision, project administration (supporting), writing -- review \& editing.

\section*{Additional information}

\subsection*{Correspondence}
Correspondence and requests for materials should be addressed to András Biricz or Péter Pollner.

\subsection*{Equal contribution}
Péter Pollner and István Csabai contributed equally to this work.

\subsection*{Competing interests}

András Biricz reports past contractual work (concluded prior to manuscript preparation) with Information in Images Ltd., directed by Michael Broderick, a company engaged in the commercial development of microscopy devices. The company provided partial support during data acquisition and computational infrastructure development and may potentially benefit from advances in automated microscopy and pollen dataset generation. The study design, data analysis, and manuscript preparation were conducted independently of any commercial interest. All other authors declare no competing interests.

\bibliography{references}

\appendix
\section*{Supplementary Information}

\renewcommand{\thetable}{S\arabic{table}}
\renewcommand{\theHtable}{S\arabic{table}}
\setcounter{table}{0}

\renewcommand{\thefigure}{S\arabic{figure}}
\renewcommand{\theHfigure}{S\arabic{figure}}
\setcounter{figure}{0}

\subsection*{Appendix A: representation visualisation}

\begin{figure}[ht]
\centering
\includegraphics[width=\textwidth]{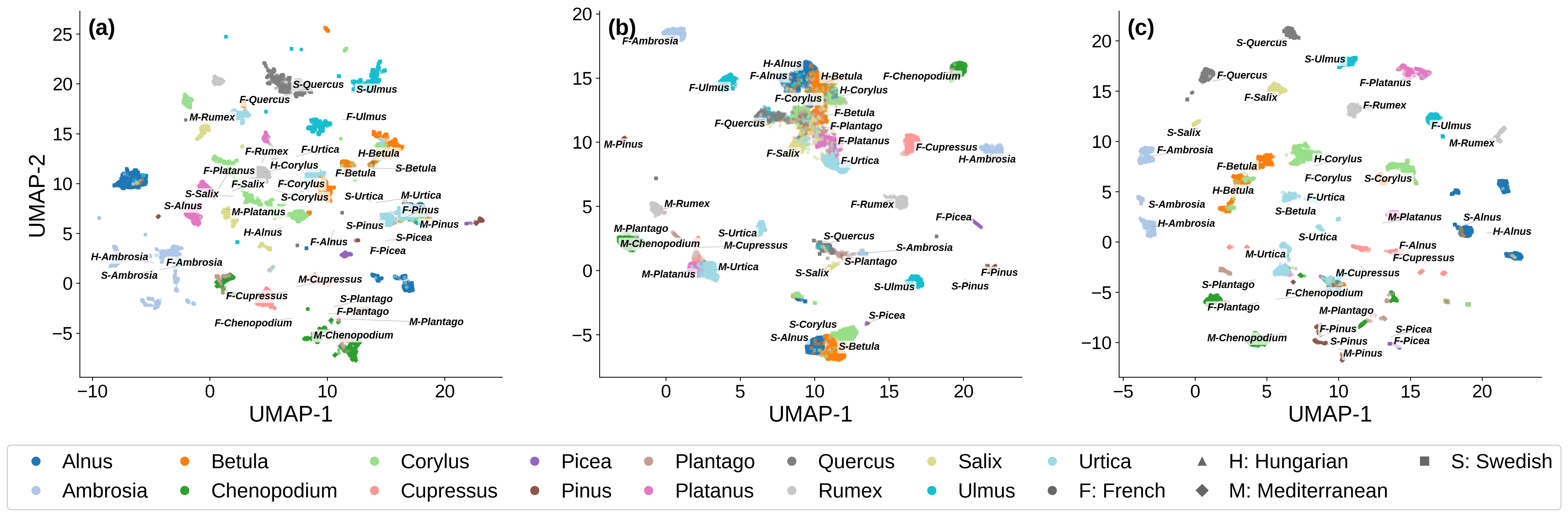}
\caption{Qualitative UMAP projections of pollen-grain embeddings across three modality spaces. a, SBERT embeddings of Gemma4 captions. b, original frozen LVD-ViT-S image features. c, Combined multimodal distances from late fusion ($\alpha=0.5$). Points are coloured by genus-level target for the 15 cross-regional retrieval targets, and point shape indicates geographic origin (French, circle; Hungarian, triangle; Mediterranean, diamond; Swedish, square). The projections were generated from 25,953 validation-split grains using cosine distance for the caption and image embeddings, precomputed distances for the combined space, 15 neighbours, a minimum distance of 0.1, and seed\,=\,42. The plots are intended as qualitative visualisations of the retrieval embedding spaces, not as quantitative evidence of cluster separability.}
\label{fig:umap}
\end{figure}

UMAP projections were used as a qualitative visual check of the embedding structure underlying the retrieval experiments. Three spaces were visualised for the 15 cross-regional genera used in retrieval: SBERT embeddings of Gemma4 captions, ViT image embeddings, and a combined late-fusion distance. Caption embeddings were computed from the generated Gemma4 captions as used for text retrieval, and image embeddings were computed from the corresponding grain crops; no additional manual cleaning, outlier removal, or caption-content filtering was applied specifically for the UMAP visualisation. Points were sampled from the validation split and balanced across genus and geographic origin, with at most 1,000 grains selected for each genus--origin combination; the final projection contained 25,953 grains. UMAP was run with 15 neighbours and a minimum distance of 0.1, using cosine distance for text and image embeddings and precomputed distances for the combined space (Supplementary Fig.~\ref{fig:umap}).

The caption embedding projection formed compact genus-level clusters with relatively limited separation by geographic origin. The image embedding projection also showed genus-level structure, but with clearer subdivision by origin, consistent with the origin sensitivity observed in CROSS-REG image retrieval. The combined projection showed an intermediate pattern. Because UMAP is a qualitative dimensionality-reduction visualisation rather than an inferential test, isolated islands or local separations may reflect a mixture of biological morphology, origin effects, visual artefacts, caption wording, and projection geometry. These projections are therefore presented as descriptive support for the retrieval results rather than as standalone evidence.

\subsection*{Appendix B: data records and usage notes}

The processed data records described in this paper are deposited in Zenodo at \footnote{\url{https://zenodo.org/records/20690944}} under a CC BY 4.0 licence, in accordance with the FAIR guiding principles for scientific data management\cite{sdata201618}. The Zenodo release\cite{zenodo_deposit} is distributed as \texttt{pollen\_ai\_atlas.zip} (42.8\,GB) and contains filtered detection metadata for all 85 processed slides, all five caption sets for the 80-slide captioned release subset, expert-curated reference annotations, pre-extracted pollen grain crops, validated ground-truth pollen crops, train/validation split definitions, supporting metadata, caption anchors, query images, and the fine-tuned LVD-ViT-Small/14 checkpoint used for mining and quality filtering. Supplementary Table~\ref{tab:data_records} summarises the released records and access routes. The source whole-slide microscopy images are not included in the Zenodo release because of data-volume constraints and are available from the corresponding author on reasonable request.

\subsubsection*{Repository structure and access}

The Zenodo release is organised by dataset origin and processing stage. At top level, the deposit contains per-dataset directories for the French, Hungarian, Mediterranean, and Swedish material, together with shared annotation, split, crop, metadata, and model files. Within each dataset directory, filtered HDF5 records are stored separately from the deposited model-specific caption outputs. The five caption sets follow the same per-slide JSONL structure for Qwen2.5-VL, Qwen3-VL, Qwen3.5, Qwen3.6, and Gemma4.

\subsubsection*{Whole-slide images}

The source microscopy archive comprises 85 processed pyramidal TIFF whole-slide images organised into four dataset subfolders (French, Hungarian, Mediterranean, and Swedish). Together these files occupy 148.9 GB (138.7 GiB), which exceeds the practical size of the Zenodo package. All 85 slides were processed through mining and filtering, and 80 contributed captioned records to the main release. The TIFF files are compatible with OpenSlide\cite{Goode2013OpenSlide} and related readers for pyramidal microscopy data. Resolution metadata are stored in the TIFF headers and are accompanied by the deposited slide-level calibration file described below.

\subsubsection*{Filtered detection metadata}

Filtered detection records are deposited as per-slide HDF5 files, one file for each of the 85 processed slides. Raw mining detection files are not included in the Zenodo release. Each filtered file contains one record per retained detection, including segmentation masks, point prompts used for segmentation, crop origins in whole-slide coordinates, bounding boxes, original mining confidence values, refined confidence scores, classifier predictions, classifier probabilities, and rebuilt token-index metadata. These filtered HDF5 files provide the spatial, segmentation, and confidence metadata linking released crops and captions to their source detections for slides with deposited caption records.

\subsubsection*{Caption files}

Caption records are generated as JSONL files, with one line per pollen grain and one file per slide for each model. The Zenodo release contains five caption sets generated by Qwen2.5-VL, Qwen3-VL, Qwen3.5, Qwen3.6, and Gemma4. Gemma4 was selected as the primary VLM output for downstream multimodal analysis, while the four Qwen-family caption sets are deposited as companion corpora for model comparison and reuse. Each JSON record contains the grain identifier, taxonomic fields, spatial metadata, quality scores, morphometric measurements, the model prompt, and the generated caption text. Together these records constitute the text modality layer of the dataset.

\subsubsection*{Pollen grain image crops}

Pre-extracted pollen grain crops are deposited as PNG files organised by dataset and slide. Each crop corresponds to one filtered detection retained in the 80-slide captioned release subset and is extracted at native whole-slide resolution, centred on the detected object with context padding. Per-slide crop manifests link each crop to its caption record and filtered HDF5 entry by grain identifier and record the crop filename, whole-slide bounding box, crop origin, and crop size. These crops represent the image modality used in downstream classification, retrieval, and visualisation workflows.

\subsubsection*{Expert-curated annotations and legacy compatibility}

Expert-curated reference annotations are deposited as GeoJSON files. The primary expert-curated evaluation set (TS2) comprises one file per annotated slide from the proposal-correction workflow described in Methods. TS2 includes 80 slides used in the main quantitative evaluation (79 slides with expert-verified taxon identifications at genus level or finer and one slide identified only to the family Fabaceae, for which the genus-level label was recorded as \textit{Unknown}) and four additional slides with nominal, unverified family-level source labels released for reference. A second set, the legacy compatibility set (TS1), is also provided to support comparison with the evaluation protocol of our previous study\cite{Biricz2025}. GeoJSON features include annotation geometry, taxonomic fields, and source tags indicating whether each object originated from a retained pipeline proposal or from expert addition during correction. The files are directly loadable in QuPath\cite{Bankhead2017QuPath}.

TS1 comprises 21 slides originally annotated in COCO format on fixed 7,500$\times$7,500\,px test-region crops, with physical area varying by pixel pitch. For this release, TS1 annotations were converted to the same GeoJSON schema as TS2, with coordinates mapped back to whole-slide space. Seventeen TS1 slides carry genus-level labels used for quantitative evaluation, yielding 5,065 grains across 11 genus-level targets; the remaining four slides lack caption anchors for the taxa present and are deposited for reference only.

Because TS1 and TS2 partially reuse the same physical slides, their test regions were checked for spatial overlap. Eighteen slides are shared between the two sets. Seventeen were verified to be spatially disjoint, whereas one \textit{Alnus glutinosa} 40$\times$ slide showed an approximately 40\% overlap between the legacy and expert-defined regions. This exception is documented explicitly in the deposited metadata. The release also provides 17,318 validated ground-truth pollen crops extracted from all GeoJSON features classified as pollen in TS2 and TS1, together with per-slide and aggregate CSV manifests linking each crop to its source GeoJSON feature and whole-slide coordinates.

\subsubsection*{Train/validation splits and model checkpoint}

Train/validation partitions are provided as per-slide JSON files together with a manifest recording the split parameters. The release contains 80 training split files, 80 validation split files, and one manifest file, with all assignments defined by grain identifier. These split definitions are model-agnostic and apply equally to all five caption sets. Samples intersecting the TS1 or TS2 test regions were removed before split generation, as described in Methods.

The release also includes the fine-tuned LVD-ViT-S checkpoint used for mining feature extraction and filtering-related classification. This checkpoint is provided as a PyTorch model file.

\subsubsection*{Slide metadata and support files}

Slide-level metadata are provided in a deposited manifest, together with companion caption-anchor text files and matched query images. The manifest records the pixel-to-micrometre calibration used for size measurements, slide-level taxonomic labels, the corresponding caption-anchor text, the public caption model and captioned-grain count where captions are deposited, and the reason any slide was excluded from captioning. Together, these files provide the calibration and source-context metadata required to interpret the released records.

\subsubsection*{Data access and loading}

The released records can be accessed with standard software libraries. Caption files are provided in JSONL format and can be read line by line with standard JSON parsers. Filtered detection metadata are stored in HDF5 format and can be accessed with standard HDF5 libraries. Expert-curated annotations are provided as GeoJSON files and can be inspected either programmatically or in annotation software such as QuPath. Crop manifests preserve object identifiers, crop filenames, whole-slide coordinates, and crop geometry for both the captioned crop corpus and the ground-truth crop subset. When obtained separately from the corresponding author, the source whole-slide images can be accessed as pyramidal TIFF files with OpenSlide\cite{Goode2013OpenSlide} or related slide-image readers.

For most downstream analyses described in this paper, the released pollen grain crops, ground-truth crops, filtered HDF5 records, GeoJSON annotations, and caption JSONL files are sufficient. Access to the original whole-slide images is only required for workflows that re-extract image regions, inspect full-slide context around the deposited crops, or rerun slide-level processing.

\subsubsection*{Reuse considerations}

The released train/validation splits and the fine-tuned LVD-ViT-S checkpoint provide a starting point for supervised classification experiments on the released crops. The caption files can be used directly for text-based retrieval, multimodal retrieval, or caption-informed classification workflows. Because the dataset spans four geographic origins, multiple scanners, and heterogeneous staining and preparation conditions, it is also suitable for evaluating robustness under cross-domain transfer. The paired image--text structure further supports development of domain-adapted multimodal models for representation learning, retrieval, and caption-guided analysis.

Users who train classification models on the released crops should use the provided split definitions rather than generating new random partitions, because the deposited splits exclude detections intersecting the TS1 and TS2 test regions. For retrieval experiments, leave-origin-out evaluation provides a stricter reuse setting by excluding grains from the same geographic origin and thereby reducing trivial scanner- or preparation-specific matches.

\subsubsection*{Known limitations}

The released detection set prioritises precision over completeness. As a consequence, the captioned corpus contains a high proportion of genuine pollen detections, but not all pollen grains present in the original slides are included. The captions are machine-generated rather than written independently for each grain by a human expert, and individual records may therefore contain omissions or phrasing artefacts. Scanner systems, effective resolution, preparation protocol, and staining characteristics vary across the four source collections, and these differences should be taken into account when comparing models across origins. Class abundance is also highly imbalanced across taxa, so balanced sampling or class-weighted losses may be useful in supervised classification. Finally, five filtered slides were excluded from caption generation because they did not support stable taxonomic anchoring or final release quality; these exclusions are documented in the deposited exclusion metadata.

\clearpage
\subsection*{Appendix C: supplementary figures}

\begin{figure}[ht!]
\centering
\includegraphics[width=1.0\textwidth]{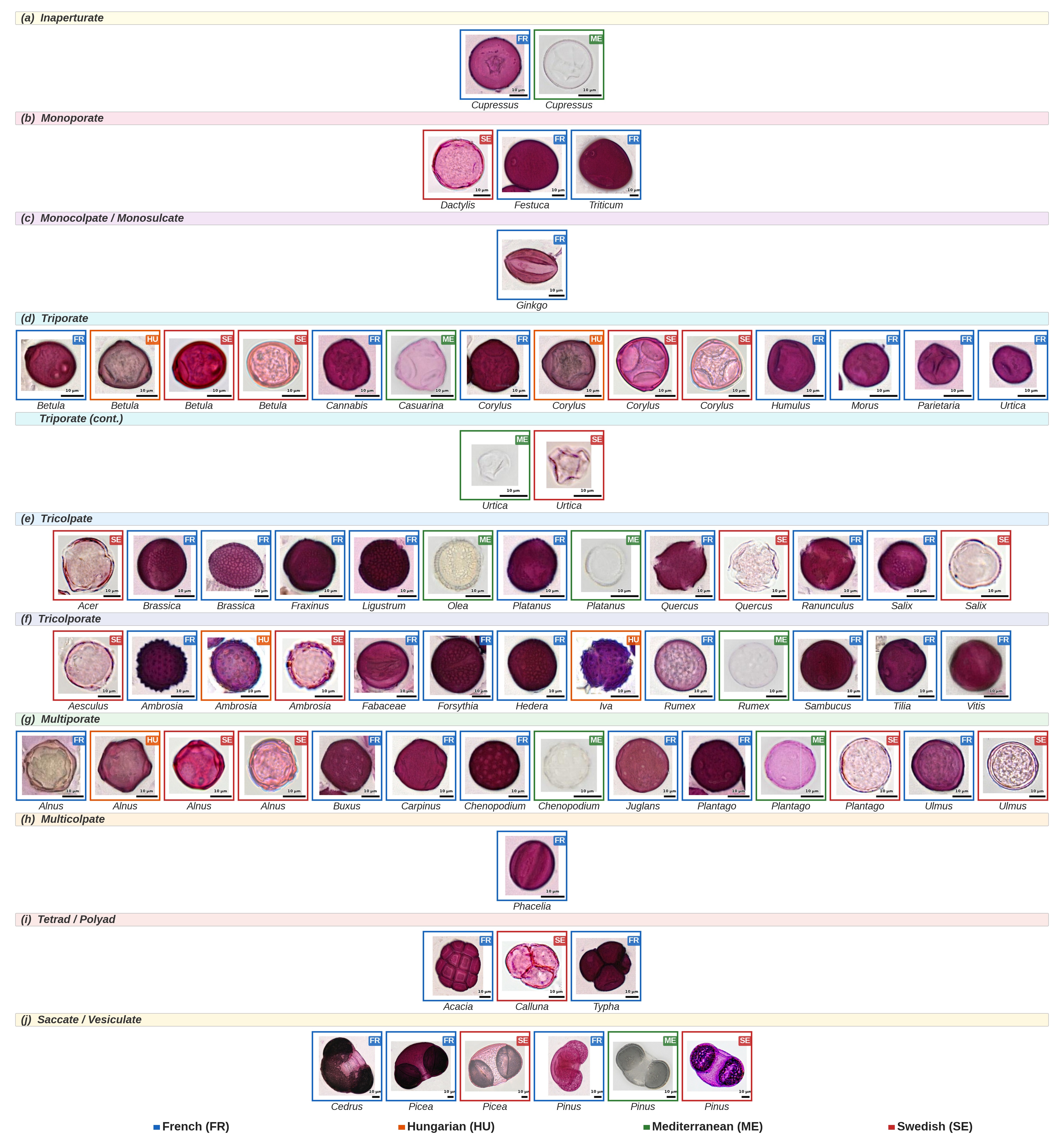}
\caption{Manual exemplar seed crops used to initialise pollen-grain mining. Exemplar crops selected from pure-species reference slides are grouped by major aperture or pollen-unit morphology in standard palynological order: (a) inaperturate, (b) monoporate, (c) monocolpate or monosulcate, (d) triporate, (e) tricolpate, (f) tricolporate, (g) multiporate, (h) multicolpate, (i) tetrad or polyad pollen units, and (j) saccate or vesiculate grains. Labels below each crop indicate the assigned taxon. Coloured borders and corner badges indicate dataset origin: French (FR, blue), Hungarian (HU, orange), Mediterranean (ME, green) and Swedish (SE, red). Repeated taxa illustrate cross-origin or slide-level variation where available. Each crop includes a 10\,\textmu m scale bar.}
\label{fig:grain_showcase}

\end{figure}

\clearpage

\subsection*{Appendix D: supplementary tables}

\begin{table}[!h]
\centering
\small
\begin{tabular}{llcrr}
\toprule
Dataset & Scanner & Objective & \textmu m/px & Slides \\
\midrule
French & 3DHistech P250 Flash III & 20× & 0.243 & 49 \\
Hungarian (20×) & 3DHistech P250 Flash III & 20× & 0.243 & 3 \\
Hungarian (40×) & 3DHistech Pannoramic 1000 & 40× & 0.121 & 2 \\
Mediterranean & Olympus VS120 & 40× & 0.170 & 11 \\
Swedish (40×) & Objective Imaging & 40× & 0.122 & 7 \\
Swedish (20×) & Objective Imaging & 20× & 0.243 & 13 \\
\midrule
\textbf{Total} & --- & --- & --- & \textbf{85} \\
\bottomrule
\end{tabular}
\caption{Whole-slide image collections used for dataset generation. For each collection or acquisition subgroup, the table reports the scanner or imaging platform, acquisition magnification, effective pixel size, and number of slides. In total, 85 slides were processed through the mining and filtering pipeline; 80 slides were retained for caption generation and included in the released dataset. Five slides were excluded before captioning because of source-taxonomy or image-quality limitations. Per-slide pixel calibration values are provided in the deposited calibration metadata.}
\label{tab:dataset_summary}
\end{table}

\begin{table}[ht!]
\centering
\small
\begin{tabular}{llrll}
\toprule
Record & Format & Count & Size & Access/location \\
\midrule
Whole-slide images & TIFF (.tif) & 85 & 148.9 GB & On request \\
Filtered detection metadata & HDF5 (.h5) & 85 & 2.7 GB & Zenodo \\
Qwen2.5-VL captions & JSONL & 80 & 7.0 GB & Zenodo \\
Qwen3-VL captions & JSONL & 80 & 6.6 GB & Zenodo \\
Qwen3.5 captions & JSONL & 80 & 6.4 GB & Zenodo \\
Qwen3.6 captions & JSONL & 80 & 6.3 GB & Zenodo \\
Gemma4 captions & JSONL & 80 & 6.4 GB & Zenodo \\
Pollen grain crops & PNG & 1,511,390 & 35.6 GB & Zenodo \\
Ground-truth pollen crops & PNG/CSV & 17,318 & 465.7 MB & Zenodo \\
Ground truth (TS2, expert) & GeoJSON & 80/84 & 8.0 MB & Zenodo \\
Ground truth (TS1, expert, legacy) & GeoJSON & 17/21 & 5.7 MB & Zenodo \\
Train/val splits & JSON & 161 & 42.7 MB & Zenodo \\
ViT-Small-LVD checkpoint & .pth (PyTorch) & 1 & 84.3 MB & Zenodo \\
Slide metadata and support files & JSON/TXT/PNG & 486 & 5.0 MB & Zenodo \\
\bottomrule
\end{tabular}
\caption{Supplementary data records and access routes. The table summarises the principal data packages associated with the Pollen AI Atlas release, including original whole-slide images, filtered detection metadata, model-specific caption files, pollen-grain crops, validated ground-truth crops, expert ground-truth annotations, train/validation splits, model checkpoints, and supporting metadata. Core release records are deposited in Zenodo, including all five model-specific caption sets. The original whole-slide images are available from the corresponding author on reasonable request. For ground-truth records, counts are reported as evaluated slides / deposited slides; slides not used in the main quantitative evaluation are retained as reference annotations where available.}
\label{tab:data_records}
\end{table}

\begin{table}[p]
\centering
\scriptsize
\setlength{\tabcolsep}{2.6pt}
\renewcommand{\arraystretch}{0.834}
\begin{tabular}{lllrrrrrrrr}
\toprule
Dataset & Family & Taxon & Filtered & GT & TP & FP & FN & Prec. & Rec. & F1 \\
\midrule
FR & Adoxaceae & Sambucus & 1,722 & 27 & 24 & 0 & 3 & 100.0\% & 88.9\% & 94.1\% \\
FR & Amaranthaceae & Chenopodium & 6,706 & 36 & 26 & 1 & 10 & 96.3\% & 72.2\% & 82.5\% \\
FR & Araliaceae & Hedera & 8,134 & 37 & 37 & 0 & 0 & 100.0\% & 100.0\% & 100.0\% \\
FR & Asteraceae & Ambrosia & 19,350 & 76 & 64 & 3 & 12 & 95.5\% & 84.2\% & 89.5\% \\
FR & Betulaceae & Alnus & 28,061 & 136 & 99 & 1 & 37 & 99.0\% & 72.8\% & 83.9\% \\
FR & Betulaceae & Betula 1 & 67,985 & 239 & 100 & 0 & 139 & 100.0\% & 41.8\% & 59.0\% \\
FR & Betulaceae & Betula 2 & 39,920 & 182 & 100 & 0 & 82 & 100.0\% & 54.9\% & 70.9\% \\
FR & Betulaceae & Carpinus & 13,830 & 60 & 51 & 0 & 9 & 100.0\% & 85.0\% & 91.9\% \\
FR & Betulaceae & Corylus 1 & 1,580 & 55 & 50 & 0 & 5 & 100.0\% & 90.9\% & 95.2\% \\
FR & Betulaceae & Corylus 2 & 6,319 & 35 & 29 & 0 & 6 & 100.0\% & 82.9\% & 90.6\% \\
FR & Boraginaceae & Phacelia & 36,477 & 149 & 100 & 0 & 49 & 100.0\% & 67.1\% & 80.3\% \\
FR & Brassicaceae & Brassica 1 & 2,758 & 43 & 31 & 1 & 12 & 96.9\% & 72.1\% & 82.7\% \\
FR & Brassicaceae & Brassica 2 & 1,066 & 74 & 55 & 0 & 19 & 100.0\% & 74.3\% & 85.3\% \\
FR & Brassicaceae & Brassica 3 & 4,725 & 28 & 24 & 0 & 4 & 100.0\% & 85.7\% & 92.3\% \\
FR & Buxaceae & Buxus & 3,054 & 31 & 20 & 0 & 11 & 100.0\% & 64.5\% & 78.4\% \\
FR & Cannabaceae & Cannabis & 11,272 & 38 & 29 & 0 & 9 & 100.0\% & 76.3\% & 86.6\% \\
FR & Cannabaceae & Humulus 1 & 2,591 & 184 & 100 & 0 & 84 & 100.0\% & 54.3\% & 70.4\% \\
FR & Cannabaceae & Humulus 2 & 3,095 & 60 & 49 & 0 & 11 & 100.0\% & 81.7\% & 89.9\% \\
FR & Cupressaceae & Cupressus & 39,453 & 174 & 100 & 0 & 74 & 100.0\% & 57.5\% & 73.0\% \\
FR & Fabaceae & Acacia & 1,283 & 19 & 13 & 0 & 6 & 100.0\% & 68.4\% & 81.3\% \\
FR & Fabaceae & Unknown & 30,737 & 117 & 100 & 0 & 17 & 100.0\% & 85.5\% & 92.2\% \\
FR & Fagaceae & Quercus & 58,321 & 219 & 100 & 0 & 119 & 100.0\% & 45.7\% & 62.7\% \\
FR & Ginkgoaceae & Ginkgo & 28,913 & 119 & 97 & 0 & 22 & 100.0\% & 81.5\% & 89.8\% \\
FR & Juglandaceae & Juglans & 2,001 & 35 & 23 & 0 & 12 & 100.0\% & 65.7\% & 79.3\% \\
FR & Malvaceae & Tilia & 14,325 & 56 & 52 & 0 & 4 & 100.0\% & 92.9\% & 96.3\% \\
FR & Moraceae & Morus & 54,767 & 153 & 100 & 0 & 53 & 100.0\% & 65.4\% & 79.1\% \\
FR & Oleaceae & Forsythia & 5,614 & 32 & 20 & 1 & 12 & 95.2\% & 62.5\% & 75.5\% \\
FR & Oleaceae & Fraxinus & 12,604 & 68 & 53 & 0 & 15 & 100.0\% & 77.9\% & 87.6\% \\
FR & Oleaceae & Ligustrum 1 & 27,930 & 64 & 63 & 0 & 1 & 100.0\% & 98.4\% & 99.2\% \\
FR & Oleaceae & Ligustrum 2 & 1,367 & 17 & 17 & 0 & 0 & 100.0\% & 100.0\% & 100.0\% \\
FR & Pinaceae & Cedrus 1 & 308 & 16 & 16 & 0 & 0 & 100.0\% & 100.0\% & 100.0\% \\
FR & Pinaceae & Cedrus 2 & 1,468 & 20 & 17 & 0 & 3 & 100.0\% & 85.0\% & 91.9\% \\
FR & Pinaceae & Picea & 1,938 & 22 & 15 & 0 & 7 & 100.0\% & 68.2\% & 81.1\% \\
FR & Pinaceae & Pinus 1 & 566 & 13 & 11 & 0 & 2 & 100.0\% & 84.6\% & 91.7\% \\
FR & Pinaceae & Pinus 2 & 1,349 & 20 & 17 & 0 & 3 & 100.0\% & 85.0\% & 91.9\% \\
FR & Plantaginaceae & Plantago & 1,033 & 22 & 13 & 0 & 9 & 100.0\% & 59.1\% & 74.3\% \\
FR & Platanaceae & Platanus & 7,334 & 37 & 30 & 0 & 7 & 100.0\% & 81.1\% & 89.6\% \\
FR & Poaceae & Festuca & 1,457 & 25 & 23 & 0 & 2 & 100.0\% & 92.0\% & 95.8\% \\
FR & Poaceae & Triticum & 1,499 & 21 & 18 & 0 & 3 & 100.0\% & 85.7\% & 92.3\% \\
FR & Polygonaceae & Rumex & 153,482 & 755 & 100 & 0 & 655 & 100.0\% & 13.2\% & 23.4\% \\
FR & Ranunculaceae & Ranunculus & 13,644 & 74 & 49 & 2 & 25 & 96.1\% & 66.2\% & 78.4\% \\
FR & Salicaceae & Salix & 47,478 & 181 & 100 & 0 & 81 & 100.0\% & 55.2\% & 71.2\% \\
FR & Typhaceae & Typha & 8,140 & 60 & 49 & 0 & 11 & 100.0\% & 81.7\% & 89.9\% \\
FR & Ulmaceae & Ulmus & 82,384 & 325 & 100 & 0 & 225 & 100.0\% & 30.8\% & 47.1\% \\
FR & Urticaceae & Parietaria & 32,726 & 189 & 100 & 0 & 89 & 100.0\% & 52.9\% & 69.2\% \\
FR & Urticaceae & Urtica & 38,876 & 91 & 83 & 0 & 8 & 100.0\% & 91.2\% & 95.4\% \\
FR & Vitaceae & Vitis & 37,731 & 144 & 100 & 0 & 44 & 100.0\% & 69.4\% & 82.0\% \\
\midrule
HU & Asteraceae & Ambrosia & 23,815 & 71 & 66 & 0 & 5 & 100.0\% & 93.0\% & 96.4\% \\
HU & Asteraceae & Iva & 21,507 & 50 & 48 & 0 & 2 & 100.0\% & 96.0\% & 98.0\% \\
HU & Betulaceae & Alnus & 293,274 & 1,003 & 100 & 0 & 903 & 100.0\% & 10.0\% & 18.1\% \\
HU & Betulaceae & Betula & 43,195 & 122 & 100 & 0 & 22 & 100.0\% & 82.0\% & 90.1\% \\
HU & Betulaceae & Corylus & 48,428 & 135 & 100 & 0 & 35 & 100.0\% & 74.1\% & 85.1\% \\
\midrule
ME & Amaranthaceae & Chenopodium & 9,310 & 214 & 100 & 0 & 114 & 100.0\% & 46.7\% & 63.7\% \\
ME & Casuarinaceae & Casuarina & 1,805 & 17 & 15 & 2 & 2 & 88.2\% & 88.2\% & 88.2\% \\
ME & Cupressaceae & Cupressus & 2,615 & 55 & 48 & 0 & 7 & 100.0\% & 87.3\% & 93.2\% \\
ME & Oleaceae & Olea & 1,959 & 42 & 40 & 1 & 2 & 97.6\% & 95.2\% & 96.4\% \\
ME & Pinaceae & Pinus & 869 & 50 & 21 & 0 & 29 & 100.0\% & 42.0\% & 59.2\% \\
ME & Plantaginaceae & Plantago & 552 & 11 & 10 & 0 & 1 & 100.0\% & 90.9\% & 95.2\% \\
ME & Platanaceae & Platanus & 4,873 & 482 & 100 & 0 & 382 & 100.0\% & 20.7\% & 34.4\% \\
ME & Polygonaceae & Rumex & 3,548 & 80 & 77 & 1 & 3 & 98.7\% & 96.2\% & 97.5\% \\
ME & Urticaceae & Urtica & 5,716 & 299 & 99 & 1 & 200 & 99.0\% & 33.1\% & 49.6\% \\
\midrule
SE & Asteraceae & Ambrosia & 1,669 & 157 & 100 & 0 & 57 & 100.0\% & 63.7\% & 77.8\% \\
SE & Betulaceae & Alnus 1 & 12,038 & 42 & 37 & 0 & 5 & 100.0\% & 88.1\% & 93.7\% \\
SE & Betulaceae & Alnus 2 & 618 & 20 & 18 & 0 & 2 & 100.0\% & 90.0\% & 94.7\% \\
SE & Betulaceae & Betula 1 & 8,661 & 24 & 22 & 0 & 2 & 100.0\% & 91.7\% & 95.7\% \\
SE & Betulaceae & Betula 2 & 10,407 & 41 & 39 & 0 & 2 & 100.0\% & 95.1\% & 97.5\% \\
SE & Betulaceae & Betula 3 & 1,574 & 79 & 46 & 0 & 33 & 100.0\% & 58.2\% & 73.6\% \\
SE & Betulaceae & Corylus 1 & 24,824 & 61 & 51 & 0 & 10 & 100.0\% & 83.6\% & 91.1\% \\
SE & Betulaceae & Corylus 2 & 2,551 & 74 & 72 & 0 & 2 & 100.0\% & 97.3\% & 98.6\% \\
SE & Ericaceae & Calluna & 107 & 13 & 11 & 4 & 2 & 73.3\% & 84.6\% & 78.6\% \\
SE & Fagaceae & Quercus & 5,065 & 317 & 100 & 0 & 217 & 100.0\% & 31.5\% & 48.0\% \\
SE & Pinaceae & Picea & 71 & 12 & 11 & 0 & 1 & 100.0\% & 91.7\% & 95.7\% \\
SE & Pinaceae & Pinus & 42 & 13 & 10 & 0 & 3 & 100.0\% & 76.9\% & 87.0\% \\
SE & Plantaginaceae & Plantago & 2,109 & 202 & 100 & 0 & 102 & 100.0\% & 49.5\% & 66.2\% \\
SE & Poaceae & Dactylis & 71 & 12 & 9 & 2 & 3 & 81.8\% & 75.0\% & 78.3\% \\
SE & Salicaceae & Salix & 1,474 & 343 & 100 & 0 & 243 & 100.0\% & 29.2\% & 45.1\% \\
SE & Sapindaceae & Acer & 341 & 36 & 26 & 0 & 10 & 100.0\% & 72.2\% & 83.9\% \\
SE & Sapindaceae & Aesculus & 903 & 98 & 63 & 0 & 35 & 100.0\% & 64.3\% & 78.3\% \\
SE & Ulmaceae & Ulmus & 5,636 & 268 & 100 & 0 & 168 & 100.0\% & 37.3\% & 54.3\% \\
SE & Urticaceae & Urtica & 4,390 & 1,314 & 100 & 0 & 1,214 & 100.0\% & 7.6\% & 14.1\% \\
\bottomrule
\end{tabular}
\caption{Per-slide filtered-detection counts and detection metrics for the 80 evaluated slides. Each row corresponds to one slide. The Filtered column reports the full-slide number of quality-filtered detections retained for release and caption generation. Ground-truth grain count (GT), true positives (TP), false positives (FP), false negatives (FN), precision, recall, and F1 are computed within the corresponding expert-curated test region. Dataset abbreviations are FR (French), HU (Hungarian), ME (Mediterranean), and SE (Swedish). The taxon column gives the slide-level display label used for evaluation; when multiple slides from the same dataset share a taxon label, numeric suffixes distinguish slide instances and do not imply separate taxonomic categories.}
\label{tab:detection_per_slide}
\end{table}
\end{document}